\documentclass[11pt,a4paper]{article}
\usepackage[margin=1in]{geometry}
\usepackage[english]{babel}

\usepackage{graphicx}
\usepackage{floatrow}
\usepackage{sidecap}

\usepackage{xcolor}

\usepackage{amsmath}
\usepackage{amssymb}
\usepackage{bm}
\usepackage{nicefrac}

\usepackage{rotating, tabularx}
\usepackage{array}
\usepackage{booktabs} \usepackage{multirow}
\usepackage{makecell}
\usepackage{adjustbox}

\usepackage[utf8]{inputenc}

\usepackage{textcomp,marvosym}

\usepackage{hyperref}
\hypersetup{hidelinks}

\usepackage{authblk}

\usepackage[nohyperlinks,nolist]{acronym}
\acrodef{AI}[AI]{artificial intelligence}
\acrodef{ANN}[ANN]{artificial neural network}
\acrodef{DNN}[DNN]{deep neural network}
\acrodef{SNN}[SNN]{spiking neural network}
\acrodef{CNN}[CNN]{convolutional neural network}
\acrodef{GPU}[GPU]{graphics processing unit}
\acrodef{CPU}[CPU]{central processing unit}
\acrodef{SGD}[SGD]{stochastic gradient descent}
\acrodef{LIF}[LIF]{leaky integrate-and-fire}
\acrodef{PSP}[PSP]{postsynaptic potential}
\acrodef{SHD}[SHD]{Spiking Heidelberg Digits}
\acrodef{ResNet}[ResNet]{residual network}
\acrodef{SG}[SG]{surrogate gradient}
\acrodef{CSNN}[CSNN]{convolutional spiking neural network}
\acrodef{ReLU}[ReLU]{rectified linear unit}
\acrodef{BPTT}[BPTT]{back-propagation through time}

\usepackage[natbib=true,sorting=none,giveninits=true,
doi=true,url=false,maxcitenames=1,maxbibnames=42, 
citestyle=numeric-comp]{biblatex}
\usepackage{csquotes}

\DeclareNameFormat{default}{\nameparts{#1}\usebibmacro{name:family-given}
     {\namepartfamily}
     {\namepartgiveni}
     {\namepartprefix}
     {\namepartsuffix}\usebibmacro{name:andothers}}

\usepackage{threeparttablex}

\usepackage[aboveskip=1pt,labelfont=bf,labelsep=period,singlelinecheck=off]
{caption}

\title{\huge Fluctuation-driven initialization for spiking\\neural network training}
\author[1,2,\Yinyang]{Julian Rossbroich}
\author[1,\Yinyang]{Julia Gygax}
\author[1,*]{Friedemann Zenke}

\affil[1]{\small Friedrich Miescher Institute for Biomedical Research, Basel,
Switzerland}
\affil[2]{\small Faculty of Science, University of Basel, Switzerland \vspace{1em}}
\affil[\Yinyang]{\small These authors contributed equally to this work.}
\affil[*]{\small Corresponding author: \href{mailto:friedemann.zenke@fmi.ch}{friedemann.zenke@fmi.ch}}

\date{} 

\begin{document}

\maketitle

\begin{abstract}
\Acp{SNN} underlie low-power, fault-tolerant information processing in the brain and could constitute a power-efficient alternative to conventional deep neural networks when implemented on suitable neuromorphic hardware accelerators. 
However, instantiating \acp{SNN} that solve complex computational tasks in-silico remains a significant challenge. 
\Ac{SG} techniques have emerged as a standard solution for training \acp{SNN} end-to-end. 
Still, their success depends on synaptic weight
initialization, similar to conventional \acp{ANN}.
Yet, unlike in the case of \acp{ANN}, it remains elusive what 
constitutes a good initial state for an \ac{SNN}.
Here, we develop a general initialization strategy for \acp{SNN} inspired by the fluctuation-driven regime commonly observed in the brain. 
Specifically, we derive practical solutions for data-dependent weight initialization that ensure fluctuation-driven firing in the widely used \ac{LIF} neurons. 
We empirically show that \acp{SNN} initialized following our strategy exhibit superior learning performance when trained with \acp{SG}. 
These findings generalize across several datasets and \ac{SNN} architectures, including fully connected, deep convolutional, recurrent, 
and more biologically plausible \acp{SNN} obeying Dale's law.
Thus fluctuation-driven initialization provides a practical, versatile, and easy-to-implement strategy for improving \ac{SNN} training performance on diverse tasks in neuromorphic engineering and computational neuroscience.
\end{abstract}

\section*{Introduction}
Spiking neurons communicate through discrete action potentials, or spikes, thereby enabling energy efficient and reliable information processing in neurobiological and neuromorphic systems \citep{sterling_principles_2017, indiveri_neuromorphic_2011}.
Before using an \ac{SNN} for any application, their connections need to be task-optimized. 
In conventional \acp{ANN} this step is accomplished through direct end-to-end optimization using back-propagation in combination with suitable parameter initialization \citep{poole_exponential_2016}.
However, the lack of smooth derivatives of neuronal spiking dynamics precludes using gradient-based optimization in \acp{SNN}.
One increasingly common approach to overcome this issue is \ac{SG} learning \citep{Hunsberger2015-kf,Zenke2021-zg, Neftci2019-ie} which relies on continuous relaxations of the actual gradients for parameter updates.  
While \acp{SG} are a powerful tool for building functional \ac{SNN} models, they can be adversely affected by poor initial parameter choices. 
In deep \acp{ANN}, suboptimal weight initialization can lead to vanishing or exploding gradients \citep{Hochreiter1991-uo,Hochreiter1997-hj}, thereby creating a major impediment for their use. 
Optimal weight initialization \citep{Glorot2010-dj, He2015-kv, Mishkin2015-zd} combined with suitable architectural choices such as skip connections \citep{Srivastava2015-yw,He2015-kv} largely avoid this issue in \acp{ANN}.
Similarly, the problem of vanishing gradients has been suggested to affect deep \acp{SNN} \citep{Lee2016,Ledinauskas2020-jk}.
However, we still lack a principled strategy for \ac{SNN} initialization.

Here, we close this gap by introducing a practical weight initialization strategy for \acp{SNN}. 
Specifically, we draw inspiration from neurobiology, where neuronal dynamics commonly exhibit fluctuation-driven firing \citep{Tiesinga2000-fc, Kuhn2004-yi}. 
Since neurons in the fluctuation-driven regime are more sensitive to small changes in the input \citep{Petersen2016-kg} and thus also to changes of their synaptic weights, we hypothesized that this regime could be advantageous for subsequent \ac{SG} learning. 
In the following, we develop a general, yet simple initialization theory for \acp{SNN} consisting of \ac{LIF} neurons, and empirically demonstrate its effectiveness for task-optimizing \acp{SNN} using \ac{SG} techniques.

\section*{Results}

Neurons in biological \acp{SNN} commonly exhibit irregular and asynchronous firing dynamics \citep{Tiesinga2000-fc, Kuhn2004-yi, vogels_neural_2005}.
Such dynamics can often be attributed to large sub-threshold fluctuations that can naturally arise through excitatory-inhibitory balance commonly observed in neurobiology \citep{brunel_dynamics_2000, vogels_neural_2005}.
To test  whether this fluctuation-driven regime could constitute a suitable initial state for subsequent learning, we proceeded in two steps. 
First, we derived a set of compact analytical expressions that link the initial synaptic weight distribution with the magnitude of sub-threshold fluctuations.
Second, we numerically tested whether initializing \acp{SNN} in the fluctuation-driven regime would allow to rapidly train these networks to high accuracy using \acp{SG}.

\begin{figure}[tbh]
\includegraphics[width=1\textwidth]{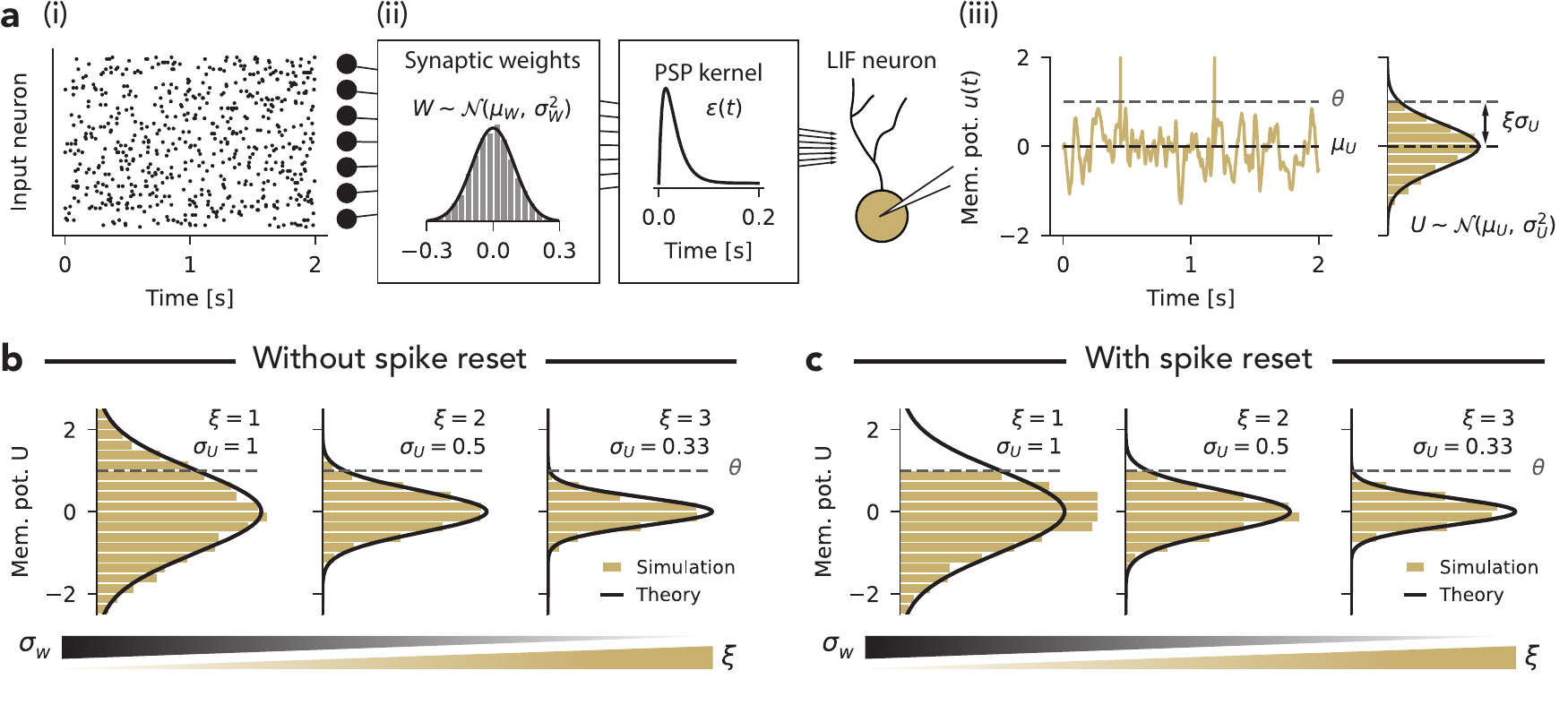}
    \caption{
    \textbf{Parameterization of fluctuation-driven spiking serves as
an initialization strategy for \acp{SNN}.} 
\textbf{(a)} Incoming presynaptic Poisson spike trains (i) are weighted by
synaptic strengths $w_j$ and filtered through a \ac{PSP} kernel $\epsilon(t)$
(ii) to yield membrane fluctuations $u(t)$ in a postsynaptic neuron (iii). 
In the fluctuation-driven regime, the membrane potential crosses the firing
threshold $\theta$ stochastically, resulting in irregular output spike trains.
Because the magnitude of membrane potential fluctuations, $\sigma_U$, is
determined by the parameters of the presynaptic weight distribution, $\mu_W$
and $\sigma_W$, synaptic weights can be initialized from a target value for the
fluctuation magnitude.
\textbf{(b)} Expected and observed distributions of the membrane potential
without considering spike-reset dynamics for different target fluctuation
strengths expressed in terms of $\sigma_U$ and $\xi$. 
\textbf{(c)} As panel (b), but considering the
spike reset dynamics in the numerical simulations.
}
\label{fig:theory}
\end{figure}

To arrive at analytical expressions, we note that there are primarily three factors that contribute to the membrane potential fluctuations (Fig.~\ref{fig:theory}a).
These are, first, the number and firing statistics of the input neurons, second, the synaptic weight distribution and third, the postsynaptic and neuronal parameters that govern temporal integration of the inputs. 
For simplicity, we assume that the presynaptic input arrives from a homogeneous population of independent Poisson neurons and that the initial weight distribution is given by a Gaussian.
Further, we limited our derivation to current-based \ac{LIF} neurons, which are commonly used in \ac{SNN} models.

To derive an expression that links the synaptic weight distribution to the
fluctuation magnitude, we consider a current-based \ac{LIF} neuron with
membrane potential $U$, whose
sub-threshold dynamics are given as the 
sum of weighted filtered presynaptic spike trains $S_j$:
\begin{equation}
	U(t) = \sum_j w_{j} \epsilon \ast S_j(t)~,
\end{equation}
where  $S_j = \sum_k \delta(t-t_j^k)$ denotes the output spike train of the presynaptic
neuron $j$ with firing times $t_j^k$ and $\ast$ is a temporal convolution of the spike train $S_j(t)$ with
$\epsilon$, a linear filter kernel with the shape of an evoked \ac{PSP}. 
Specifically, we assume a synaptic model with exponentially decaying currents
and, therefore, the shape of $\epsilon$ is fully characterized by the synaptic
and membrane time constants $\tau_{\text{syn}}$ and $\tau_{\text{mem}}$
(Methods and Supplementary Material \ref{sup:kernel}).
Since we have many statistically independent inputs, the Central Limit Theorem guarantees
that $U$ approaches a normal distribution which is fully specified by its mean
$\mu_U$ and variance $\sigma_U^2$.
Further assuming that presynaptic spikes are generated by homogeneous
Poisson processes with associated firing rates $\nu_j = \left\langle S_j
\right\rangle$,  
yields the following expressions for the mean and the variance
\begin{eqnarray} 
\mu_U &\equiv& \left\langle U \right\rangle = \sum_j w_{j} \nu_j
\int_{-\infty}^\infty \epsilon(s) ds = \sum_j
w_{j} \nu_j \bar\epsilon
\label{eq:main-mu-u} \\
\sigma^2_{U} &\equiv& \left\langle U^2 \right\rangle -\mu_U^2 =\sum_j w_{j}^2
\nu_j \int_{-\infty}^\infty \epsilon(s)^2 ds
	= \sum_j w_{j}^2 \nu_j \hat{\epsilon} ~,
	\label{eq:main-sigma-u}
\end{eqnarray}
in which $\bar \epsilon$ and $\hat \epsilon$ correspond to definite integrals
of the filter kernel and squared filter kernel respectively which can be obtained analytically for many common neuron
models (Supplementary Material \ref{sup:kernel}).
For $n$~inputs with equal firing rates $\nu_j=\nu$
and independently
drawn normally distributed weights $W \sim \mathcal{N} \left(\mu_W,
\sigma_W^2\right)$, the above expressions further simplify to
\begin{eqnarray} 
\mu_U &=& n \mu_W\nu\bar\epsilon
\label{eq:mu_u} 
\\
\sigma^2_{U}&=& n(\sigma_W^2 + \mu_W^2)\nu\hat\epsilon ~.
\label{eq:sigma_u}
\end{eqnarray}
Finally, rewriting Equations~\eqref{eq:mu_u} and~\eqref{eq:sigma_u} yields the
desired expressions linking the synaptic weight distribution with the magnitude
of the membrane potential fluctuations:
\begin{eqnarray}
\mu_W &=& \frac{\mu_U}{n \nu \bar\epsilon}
\label{eq:wbar} 
\\
\sigma^2_W &=& \frac{\sigma_U^2}{n \nu \hat\epsilon} - \mu_W^2
\nonumber
\\
&=& \frac{1}{n \nu \hat\epsilon} \left(\frac{\theta - \mu_U}{\xi} \right)^2 - \mu_W^2 ~.
\label{eq:wsigma}
\end{eqnarray}

For a neuron to be in the fluctuation-driven regime requires the bulk of the
Gaussian distribution has to lie below the firing threshold
(Fig.~\ref{fig:theory}b).
At the same time we require a non-vanishing probability to cross threshold to
ensure some baseline levels of spiking activity.
To formalize these requirements, we introduced the target parameter $\xi$ as 
\begin{equation} \label{eq:xi}
    \xi \equiv \frac{\theta-\mu_U}{\sigma_U}
\end{equation} 
which describes the distance between the mean membrane potential $\mu_U$ 
and the spike threshold $\theta$ in units of the standard deviation $\sigma_U$
(Fig.~\ref{fig:theory}a and Supplementary Fig.~\ref{sfig:theory-mu}).
To satisfy the above requirements, $\xi$ should be on the order of one.
Concretely, we consider the range $1 \leq \xi \leq 3$ (Fig.~\ref{fig:theory}b).
For zero-mean weight distributions this directly translates into 
a desired fluctuation amplitude range $\frac{1}{3} \leq \sigma_U \leq 1$,
which, given the above assumptions, is achieved by
\begin{equation}
\sigma^2_W = \frac{\sigma_U^2}{n \nu \hat\epsilon} 
\label{eq:wsigma-simple}
\end{equation}
at initialization.

The expressions are based on a no-spiking assumption.
Hence, we expect systematic deviations from the derived membrane potential
distribution in
the presence of spiking.
However, for small sub-threshold fluctuations ($\sigma_U \ll 1$),
the systematic contribution of the spike reset becomes negligible
(Fig.~\ref{fig:theory}c).
Exact membrane potential distributions that take into consideration
spike reset dynamics could be obtained using the Fokker-Planck equation \cite{Amit1997-ds,Gerstner2014-ke}, however such an approach does not yield compact analytic expressions and is, thus, less practical for our purposes.

\begin{figure}[tbh]
    \includegraphics[width=1\textwidth]{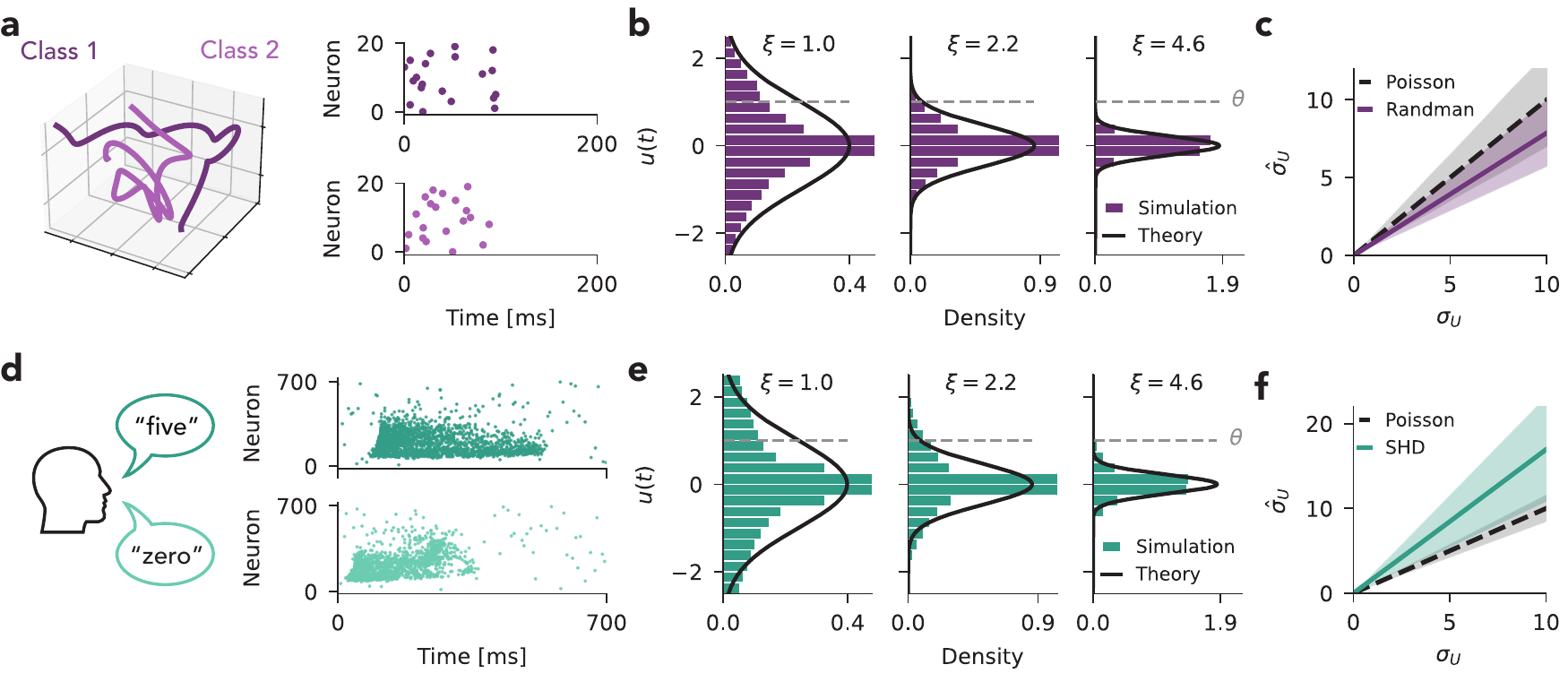}
    \caption{\textbf{Real-world datasets induce small systematic biases in
fluctuation strength at initialization.}
\textbf{(a)} Two one-dimensional example manifolds from the Randman dataset,
embedded into a three-dimensional space (left) and example spike raster plots
corresponding to a sample from each class (right).
\textbf{(b)} Theoretically expected distribution and numerically obtained density histogram of the membrane potential of a single neuron without spike reset in response to the Randman dataset. Because of large peaks at $u(t)=0$, the x-axes in the first and middle panels have been truncated to 45\% and 80\% of their maximum, respectively. 
\textbf{(c)} Numerically observed $\hat \sigma_U$ as a function of the
target $\sigma_U$ for the Randman dataset. The expected relationship
corresponds to homogeneous and independent Poisson neurons. Shaded regions
indicate standard deviation across neurons.
\textbf{(d)} Two spike rasters that correspond to two example inputs from the
SHD dataset. Input spikes are obtained by filtering recordings of spoken
digits with a biologically inspired cochlear model \citep{cramer_heidelberg_2020}.
\textbf{(e)} As panel (b), for the SHD dataset. X-axes in the first and middle panels have been truncated to 58\% and 90\% of their maximum, respectively. 
\textbf{(f)} As panel (c), for the SHD dataset.
}
\label{fig:valid}
\end{figure}

Because Eqs.~\eqref{eq:mu_u} and~\eqref{eq:sigma_u} are based on an
independence assumption that is violated by real-world data, we expected
further deviations in numerical simulations with real-world data.
To quantify the magnitude of these deviations, we compared the predictions of
Eqs.~\eqref{eq:mu_u} and~\eqref{eq:sigma_u} with observed membrane potential
fluctuations in a single \ac{LIF} neuron exposed to inputs from two realistic
datasets.
For simplicity, we assumed a zero-mean weight distribution and used
Eq.~\eqref{eq:wsigma-simple} to obtain its standard deviation for different
target fluctuation magnitudes $\sigma_U$.

First, we considered a synthetic classification dataset based on random manifolds \citep{Zenke2021-zg} (Randman; see Methods) with $n_{\mathrm{Randman}}=20$ input neurons and 10 classes in which spike times belonging to the same class are drawn from a smooth random manifold (Fig.~\ref{fig:valid}a) all the while different classes
correspond to different manifolds.
For each input pattern, each neuron fires precisely one spike during a 100\,ms
interval.
Each 100\,ms input interval was followed by 100\,ms of inactivity in the input
layer to allow for a propagation delay in the hidden layer
(Fig.~\ref{fig:valid}a).
We then recorded the membrane potential distribution and found, 
as expected, that it deviated from a Gaussian (Fig.~\ref{fig:valid}b), due to the temporal
non-stationarity and structure in the real-world data.
Next, we measured the observed membrane potential fluctuations $\hat \sigma_U$
for varying target values of $\sigma_U$ (Fig.~\ref{fig:valid}c).
We found that $\hat \sigma_U$ was systematically smaller than $\sigma_U$.
However, the magnitude of bias was comparable to the expected variability in
the case of Poisson inputs (see Methods; Supplementary Material
\ref{sup:popvar}).

Next, we considered the \ac{SHD} speech dataset (Fig.~\ref{fig:valid}d), which
consists of approximately 10,000 spoken digits in German and English that have
been converted into spikes using a biologically plausible cochlear model
\citep{cramer_heidelberg_2020}.  
Importantly, \ac{SHD} has a larger number of input neurons
($n_{\mathrm{SHD}}=700$) which typically fire more than one
spike with an average input firing rate of $\nu_{\mathrm{SHD}}=15.8$\,Hz.
Again, we measured the membrane potential distribution and observed deviations
from a Gaussian (Fig.~\ref{fig:valid}e).
In contrast to the Randman data, the observed fluctuations $\hat
\sigma_U$ were systematically larger than their target $\sigma_U$ due to heavy
tails in the distribution (Fig.~\ref{fig:valid}f).
Not surprisingly, real-world data causes systematic deviations from Eqs.~\eqref{eq:mu_u}
and~\eqref{eq:sigma_u},
but these differences were on the same order as expected fluctuations due to
finite sample size of the weight and Poisson variability.
Hence, we reasoned that our simple theory provides a reasonable approximation
for initializing \acp{SNN} in the fluctuation-driven regime even when using
real-world data.

\subsection*{Initialization of shallow \acp{SNN}}

\begin{figure}[tbhp]
    \centering
    \includegraphics[width=1\textwidth]{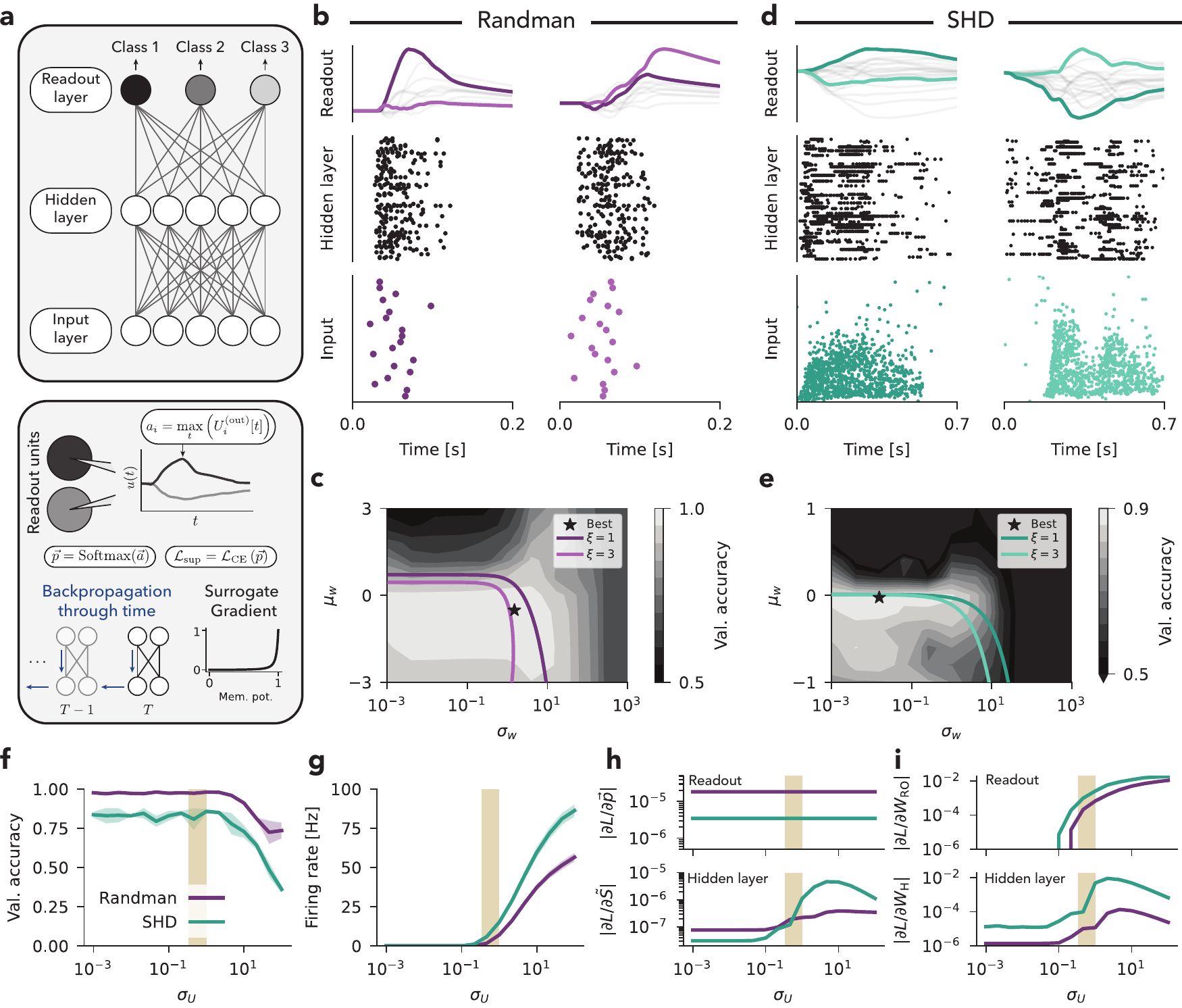}
    \caption{\textbf{Initialization in the fluctuation-driven regime results in
optimal learning performance.} 
    \textbf{(a)}~Top: Schematic of the \ac{SNN} used for
training.
    Bottom: Illustration of the learning dynamics. The supervised loss
function $\mathcal{L}_{\text{sup}}$ relies on the maximum membrane potential
over time of readout units $U_i^{(\text{out})}[t]$, to which a Softmax and
cross-entropy loss $\mathcal{L}_{\text{CE}}$ is applied. All networks were
trained by minimizing $\mathcal{L}_{\text{sup}}$ in the direction of negative
\acp{SG}, computed with \ac{BPTT}.
    \textbf{(b)}~Snapshot of network activity over time after training on the
Randman dataset. Bottom: Spike raster of input layer activity from two
different samples corresponding to two different classes is shown. Middle:
Spike raster of hidden layer activity. Top: Membrane potential of readout
units. The readout units corresponding to the two input classes are highlighted
in different shades.
    \textbf{(c)}~Heatmap showing validation accuracy after training on the
Randman dataset as a function of the parameters of the synaptic weight
distribution at initialization. 
    \textbf{(d)}~Same as in panel (b), but for a network trained on the SHD
dataset.
    \textbf{(e)}~Same as panel (c), but for the SHD dataset.
    \textbf{(f)}~Validation accuracy as a function of target fluctuation
magnitude $\sigma_U$ for initializations in the balanced state with $\mu_U =
\mu_W = 0$. 
The shaded region around the lines indicates the range of values across five random seeds. The sand-colored shaded region corresponds to our suggested target fluctuation magnitude $\frac{1}{3} \leq \sigma_U \leq 1$.
    \textbf{(g)}~Average hidden layer firing rate as a function of
$\sigma_U$.
    \textbf{(h)}~Average magnitude of \acp{SG} with respect to the 
output in the readout layer (top) and hidden layer (bottom) as a function of
$\sigma_U$.
    \textbf{(i)}~Same as panel (h), but for the average magnitude of \acp{SG} with respect to the synaptic weights.
    }
    \label{fig:shallow}
\end{figure}

We sought to evaluate whether the fluctuation-driven regime constitutes a good
initialization strategy for \ac{SNN} training.
To this end, we trained a fully connected \ac{SNN} with one hidden layer with
128~units on the Randman dataset (cf. Supplementary Tab.~\ref{stab:architecture_shallow}; Methods).
We initialized the weights using the parameters $\mu_W=0$ and $\sigma_W$ given
by our theory (Eq.~\eqref{eq:wsigma-simple} with target $\sigma_U=1$).
This choice resulted in asynchronous irregular firing activity consistent with
the fluctuation-driven regime (Supplementary Fig.~\ref{sfig:shallow}a-d). 
Subsequently, we trained the network in a supervised fashion using \acp{SG}
with previously established parameters \citep{Zenke2021-zg}, 
\ac{BPTT}, and a maximum-over-time loss defined on ten
readout units (Fig.~\ref{fig:shallow}a; Methods).
Training resulted in an \ac{SNN} that accurately solved the task ($99.6\%\pm0.0$ train \& $97.3\%\pm0.2$ test accuracy; Fig.~\ref{fig:shallow}b).

To test whether our weight initialization strategy confers an advantage over
other choices of $\mu_W$ and $\sigma_W$, we performed an extensive parameter
search and measured validation accuracy after 200~training epochs.
The network achieved the best validation accuracy when $\mu_W$ was zero or
negative and $\sigma_W$ was close to one (Fig.~\ref{fig:shallow}c), well within
our suggested regime of $1 \leq \xi \leq 3$.
Further, we found a large parameter regime that supported learning at 
close-to-optimal accuracy for $-2 \leq \mu_W \leq 0$ and $\sigma_W < 10$ 
which extends beyond the parameter regime suggested by our theory.

To test whether these results would change on a more complex task, we trained a
similar \ac{SNN} on the \ac{SHD} dataset \cite{cramer_heidelberg_2020} with weight
parameters $\mu_W=0$ and $\sigma_W^\mathrm{(SHD)}= 0.23$ as suggested by
our theory (Eq.~\eqref{eq:wsigma-simple} with target $\sigma_U=1$).
Due to differences of the number of input neurons and firing rates between the
two datasets our theory predicts $\sigma_W^\mathrm{(SHD)} \approx 10^{-1} \sigma_W^\mathrm{(Randman)}$.
After training, the network accurately classified spoken digits ($100.0\%\pm0.0$ train \& $65.5\%\pm0.7$ test accuracy; Fig.~\ref{fig:shallow}d).
As before, we performed an extensive parameter search over different
initializations and found that networks initialized in the fluctuation-driven
regime ($1 \leq \xi \leq 3$) showed close-to-optimal performance
(Fig.~\ref{fig:shallow}e, f).
Unlike in the Randman case, the parameter regime with good performance was much
smaller and tightly constrained around $\mu_W \approx 0$.
Finally, even though our initialization strategy posits that neurons be in the
fluctuation-driven regime, we observed a sizeable fraction of hidden layer
neurons with regular firing activity both before (Supplementary Fig.~\ref{sfig:shallow}e) and after learning (Fig.~\ref{fig:shallow}d). 
We found that our theory predicts these cases (Supplementary Fig.~\ref{sfig:shallow}d, h) due to the inherent variability in the sampling of
synaptic weights (Supplementary Material \ref{sup:popvar} and Supplementary Fig.~\ref{sfig:popvar}).

For both datasets, we found that initialization with $\sigma_W \ll 1$ and
$\mu_W \approx 0$ supported close-to-optimal learning.
This result surprised us, because the ensuing vanishing membrane potential
fluctuations should lead to quiescent hidden layer activity.
To check whether this is indeed the case, we initialized networks with
different target values for $\sigma_U$ and recorded their hidden layer
activity.
As expected, we found that fluctuation magnitudes $\sigma_U \ll 1$ still
supported close-to-optimal learning performance (Fig.~\ref{fig:shallow}f),
despite an absence of spikes in the hidden layer at the time of initialization
(Fig.~\ref{fig:shallow}g).

Because vanishing spiking activity should influence gradient magnitudes during backpropagation, we recorded the
magnitude of the \ac{SG} with respect to the output at the readout
and hidden layers at the time of the first training epoch.
Due to the nature of the loss function, initialization does not affect the
magnitude of the gradient in the readout layer, but can change the magnitude
of the gradient by two orders of magnitude in the hidden layer
(Fig.~\ref{fig:shallow}h). 
Consequently, the absolute magnitude of weight changes is
also amplified in the hidden layer when fluctuations are large
(Fig.~\ref{fig:shallow}i). 
Since the synaptic weight update depends on presynaptic activity,
initializations resulting in quiescent hidden layers
(Fig.~\ref{fig:shallow}g) lead to an absence of weight
updates in the readout layer (Fig.~\ref{fig:shallow}i). 
However, as long as \acp{SG} do not vanish in the first layer,
the network can recover spike propagation and therefore gradient flow during
training.
That the network is able to learn without problems in this regime may seem
surprising at first and is indeed a peculiarity of \acp{SG}.

So far, we studied strictly feed-forward \acp{SNN} without recurrent hidden
layer connections.
Recurrent \acp{SNN} typically perform better than feed-forward networks on
tasks requiring memory such as \ac{SHD} \citep{Zenke2021-zg}.
To that end, we extended our initialization strategy to networks
with recurrent connections (see Methods) and applied it to recurrent \acp{SNN}
with one hidden layer.
As in the case of feed-forward networks, we found recurrent \acp{SNN} trained
well with sufficiently small target fluctuations $\sigma_U$ (Supplementary Fig.~\ref{sfig:recurrent}a, b). 

In summary, shallow \acp{SNN} are surprisingly robust to initialization when
the absolute magnitude of the weights is small. In practice, initialization
with $\mu_U=0$ and a target fluctuation magnitude $\sigma_U \leq 1$ can be used
to achieve close-to-optimal learning performance.

\subsection*{Initialization of deep \acp{SNN}}

\begin{figure}[tb]
    \centering
    \includegraphics[width=1\textwidth]{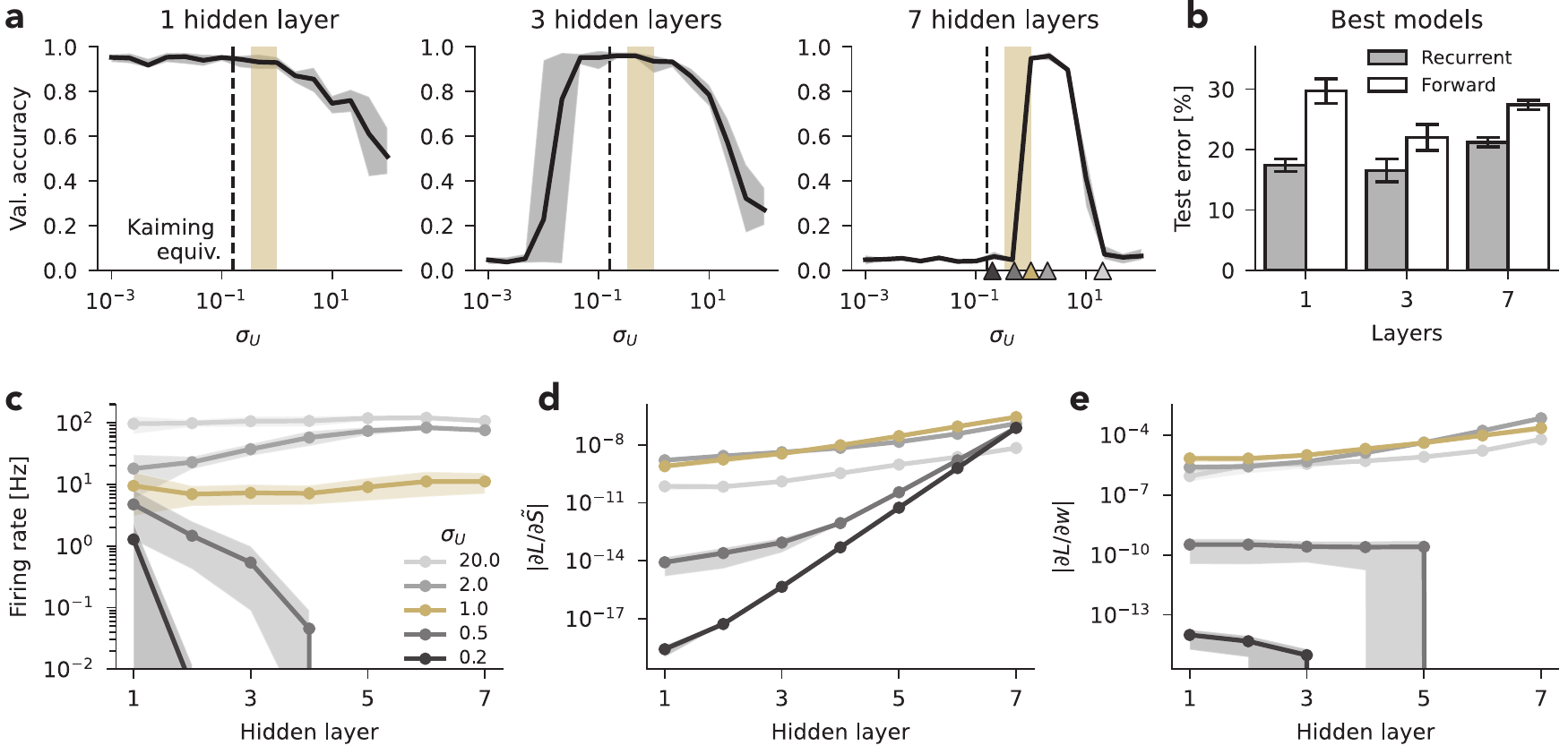}
    \caption{\textbf{Deep \acp{CSNN} are sensitive to initialization due to vanishing \acp{SG}.} 
    \textbf{(a)} Validation accuracy as a function of target fluctuation strength $\sigma_U$ for recurrent \acp{CSNN} of increasing depth. All networks were trained on the SHD dataset. The triangular markers in the right plot correspond to the values of $\sigma_U$ plotted in panels (c)-(e). The shaded region around the lines indicates the range of values across five random seeds. The sand-colored shaded region corresponds to our suggested target fluctuation magnitude $\frac{1}{3} \leq \sigma_U \leq 1$. The dashed line corresponds to Kaiming initialization.
    \textbf{(b)} Test error of the five best performing models in terms of validation accuracy, for different numbers of hidden layers and for networks with and without recurrent connections in the hidden layers.
    \textbf{(c)} Population firing rate at initialization (before training) as a function of hidden layer in a \ac{CSNN} with seven hidden layers, for different values of $\sigma_U$.
    \textbf{(d)} As panel (c), but displaying the magnitude of \acp{SG}. 
    \textbf{(e)} As panel (c), but displaying the magnitude of the synaptic weight update. When membrane potential fluctuations are so small that neurons in the previous layer do not spike, the weight update equals zero.
    }
    \label{fig:deep}
\end{figure}

We hypothesized that deep \acp{SNN} are more sensitive to initialization, as is the case with deep \acp{ANN} \citep{He2015-kv}. 
To test this hypothesis, we first extended our initialization strategy to deep and recurrent \ac{CSNN} architectures (see Methods).
We then initialized several \ac{CSNN} architectures with different numbers of recurrently connected hidden layers according to \mbox{Eqs.~(\ref{eq:muWV_nondale_rec}-\ref{eq:sigmaV_nondale_rec})} with target $\mu_U=0$ and different targets $\sigma_U$.
Subsequently, we trained the resulting networks and measured validation accuracy on held-out data.
As expected, sensitivity to the fluctuation magnitude at initialization increased with network depth (Fig.~\ref{fig:deep}a). 
As in shallow fully connected networks, \acp{CSNN} with a single hidden layer were remarkably robust to initialization and close-to-optimal training performance was achieved for $\sigma_U \leq 1$. 
In contrast, deep networks with three hidden layers performed well when the fluctuation magnitude fell into the regime \mbox{$0.1 \leq \sigma_U \leq 4$}. 
This regime was even narrower for deeper networks with seven hidden layers, which only achieved high validation accuracy for initializations in the range $1 \leq \sigma_U \leq 4$.
In summary, we observed that our theory of fluctuation-driven initialization with $\sigma_U = 1$ supported learning in deep \acp{CSNN} with up to seven hidden layers. 
For reference we repeated network training using Kaiming initialization \cite{He2015-kv}, which is commonly used for \acp{ANN}.
We found that this initialization strategy supported learning in three-layer, but not seven-layer \acp{CSNN}.

To check whether depth increases generalization performance of trained networks, we compared the test error of successfully trained \acp{CSNN} with one, three, and seven hidden layers.
We found that three-layer networks did not show better generalization performance than one-layer networks, and that seven-layer networks performed worse (Fig.~\ref{fig:deep}b).
These findings suggest that the addition of multiple hidden layers does not provide an advantage in recurrently connected \acp{SNN} on the SHD dataset.
Since recurrently connected networks can be considered as deep in time, we were wondering whether strictly feed-forward  \acp{SNN} would benefit from increasing depth.
To that end, we repeated training of deep \acp{CSNN} with corresponding layer sizes but without recurrent hidden layer connections on the SHD dataset.
Indeed, we found that deep feed-forward \acp{SNN} performed better than shallow feed-forward \acp{SNN} (Fig.~\ref{fig:deep}b and Supplementary Fig.~\ref{sfig:deep-forward}).

\paragraph{Vanishing \acp{SG} impair learning in deep \acp{SNN}.}
In deep \acp{ANN} initialization is closely related to the problem of vanishing or exploding gradients \citep{Hochreiter1991-uo,Hochreiter1997-hj}. 
We wondered whether this mechanism, i.e., vanishing or exploding \acp{SG}, prevented training in deep \acp{SNN} when $\sigma_U$ falls outside the optimal regime.
To test this idea, we initialized seven-layer \acp{CSNN} with different targets $\sigma_U$ and recorded the neuronal activity in hidden layers.
Like in shallow \acp{SNN} (Fig.~\ref{fig:shallow}g), initializations with small $\sigma_U$ led to quiescent hidden layers in deep \acp{CSNN}, which impaired the activity propagation to deeper layers (Fig.~\ref{fig:deep}c).
Specifically, in networks initialized with $\sigma_U = 0.5$, only the first four hidden layers exhibited spiking activity.
This effect was amplified in networks initialized with $\sigma_U = 0.2$, in which all but the first hidden layer were quiescent. 
In contrast, networks initialized with $\sigma_U = 2$ exhibited a strong increase of firing rates in deeper layers and initializations with $\sigma_U =20$ caused firing rates to saturate in all layers of the network.
Only initializations with $\sigma_U = 1$ led to stable activity propagation with a firing rate of $\approx10$ Hz throughout the network. 

We next investigated how impaired activity propagation influenced \ac{SG} magnitudes.
To that end, we recorded \ac{SG} magnitudes at each hidden layer during training.
In networks initialized with $\sigma_U = 0.5$ and $\sigma_U = 0.2$, in which spiking activity vanished in deep layers, each quiescent layer decreased \acp{SG} by approximately two orders of magnitude (Fig.~\ref{fig:deep}d). 
As a result, the magnitude of weight updates in early layers decreased by several orders of magnitude consistent with the numerical value of the surrogate derivative for neurons at rest ($0.023$ for $\beta=20$; see Methods).
Moreover, weight updates vanished in deeper layers, caused by the lack of presynaptic activity (Fig.~\ref{fig:deep}e).
In contrast, initializations with $\sigma_U \geq 1$ led to relatively stable \ac{SG} and weight update magnitudes across all layers (Fig.~\ref{fig:deep}d, e). 
Notably, gradients were consistently one to two orders of magnitude smaller in networks initialized with $\sigma_U = 20$ compared to networks initialized with $\sigma_U = 1$ or $\sigma_U = 2$ (Fig.~\ref{fig:deep}d, e).

In summary, the sensitivity to initialization in deep \acp{SNN} is caused by impaired activity propagation to deeper layers and associated vanishing \acp{SG}.
Empirically we found that only initializations with $\sigma_U \approx 1$ supported both propagation of sparse population activity and stable magnitudes of back-propagating \acp{SG} in deep networks.

Since the surrogate derivative used to compute \acp{SG} is to some extent freely tunable \citep{Zenke2021-zg}, one might argue that re-scaling it could provide a potential solution to vanishing \acp{SG} by ensuring stable gradient magnitudes during back-propagation (see Methods).
We tested this approach and found that a re-scaled \ac{SG} can only prevent vanishing gradients in the absence of spiking at the cost of exploding gradients when the network does exhibit spiking which emerges over training (Supplementary Fig.~\ref{sfig:surrgrad-rescaled}a-c).
In strictly feed-forward networks, we found that the gradients were less prone to exploding, hence re-scaling the \ac{SG} could potentially alleviate the problem of vanishing gradients (Supplementary Fig.~\ref{sfig:surrgrad-rescaled}d-f) and therefore increase robustness to
initialization. 
However, with increasing depth, exploding gradients would likely prevent successful training even in deep feed-forward \acp{SNN}.

Seeing that training of deep \acp{SNN} was sensitive to the magnitude of \acp{SG},
we speculated that the robustness to weight initialization we observed in three-layer \acp{CSNN} could be
attributed to the use of our optimizer with a per-parameter learning
rate during training (see Methods).
To test this idea, we trained three-layer \acp{CSNN} initialized with different $\sigma_U$
either with a smart optimizer \citep{Funk2015-xl} or with \ac{SGD} without an
optimizer.
We found that networks trained with \ac{SGD} were indeed more sensitive to the
fluctuation magnitude at initialization (Supplementary Fig.~\ref{sfig:optimizers}).
This effect was especially prominent in recurrent \acp{CSNN}.

\begin{figure}[tb]
    \centering
    \includegraphics[width=1.0\textwidth]{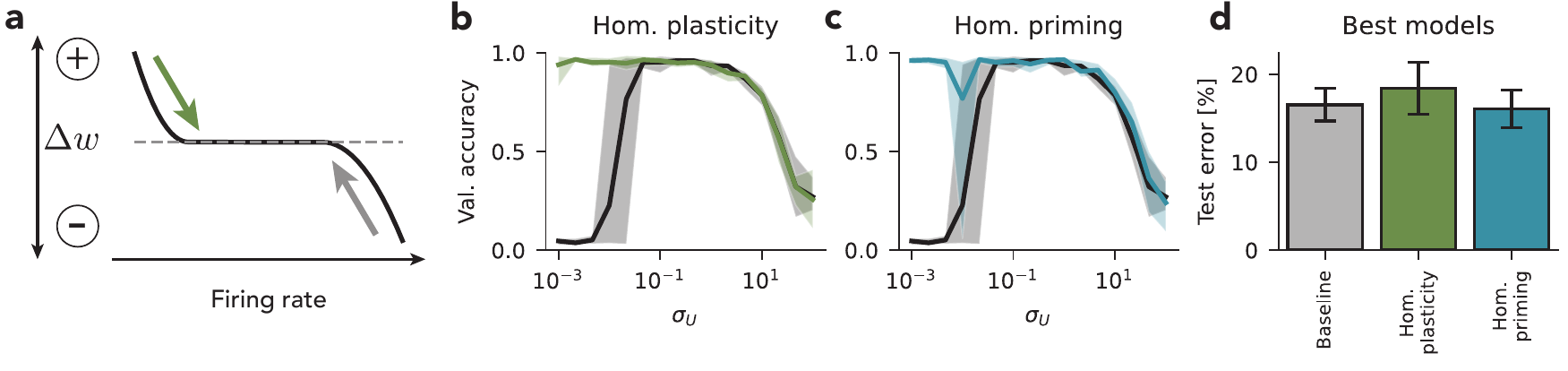}
    \caption{\textbf{Homeostatic plasticity increases the robustness to initialization in deep \acp{SNN}.} 
    \textbf{(a)} Illustration of the homeostatic activity mechanism as a firing rate regularizer. Homeostatic plasticity (green) prevents neurons from remaining silent by increasing the synaptic weights when the firing rate is low. In all our simulations, a complementary upper bound activity regularizer (grey), that acts on the population-level, prevents neurons from spiking incessantly. 
    \textbf{(b)} Validation accuracy after training a deep convolutional \ac{SNN} with three hidden layers on the \ac{SHD} dataset as a function of $\sigma_U$. The colored line corresponds to networks trained with an active homeostatic plasticity mechanism. The black line corresponds to the baseline  without homeostatic plasticity. The shaded region around the lines indicates the range of values across five random seeds. 
    \textbf{(c)} As panel (b), for networks that were primed for 10 epochs with a homeostatic plasticity mechanism prior to supervised learning. During supervised learning, the homeostatic mechanism was inactive.
    \textbf{(d)} Test error of the 5 best performing models in terms of validation accuracy for models trained with homeostatic plasticity, homeostatic priming and the baseline model. 
    }
    \label{fig:homplast}
\end{figure}

\paragraph{Homeostatic plasticity increases robustness to initialization in deep \acp{SNN}.}
Because quiescent hidden layers are closely linked to vanishing \acp{SG} and thus to preventing training in deep \acp{SNN}, a homeostatically maintained firing rate, as observed in biological neural networks \cite{turrigiano_homeostatic_2012, gjorgjieva_homeostatic_2016, zenke_hebbian_2017}, could rescue activity propagation and therefore enable training. 
To test this hypothesis, we implemented homeostatic plasticity as an additional regularization term in the loss function that sets a lower bound on the firing rate of each individual neuron \citealp{cramer_heidelberg_2020}, which penalizes quiescent neurons (Fig.~\ref{fig:homplast}a; see Methods). 
We trained three-layer recurrent \acp{CSNN} on the SHD dataset, either with or without the additional homeostatic plasticity term. 
Indeed, homeostatic plasticity rescued training performance for networks initialized with $\sigma_U \ll 1$ (Fig.~\ref{fig:homplast}b). 

Next, we investigated whether homeostatic plasticity was necessary throughout the whole training period, or whether rescuing activity propagation before supervised training would be sufficient to enable learning. 
To this end, we developed a form of dynamic initialization for \acp{SNN} involving a homeostatic priming period before training. 
During the initial priming period, initialized networks were optimized solely on the homeostatic objective to nudge the spiking activity into a regime conducive to learning. 
After priming, we removed the homeostatic objective and started the supervised training period as usual.
Like ongoing homeostatic plasticity, the homeostatic priming period was capable
of rescuing learning for initializations with $\sigma_U \ll 1$ (Fig.~\ref{fig:homplast}c).
However, in rare cases, the network did not train after successful priming and the restored spiking activity vanished during training on the supervised loss function.

We wondered whether homeostatic plasticity affected the network's generalization performance and thus compared the test error of networks trained with the proposed homeostatic mechanisms.
Neither ongoing homeostatic plasticity nor homeostatic priming had a systematic effect on the test error (Fig.~\ref{fig:homplast}d). 
Therefore, we concluded that both biologically inspired homeostatic plasticity and homeostatic priming are effective strategies to increase the robustness towards initialization in deep \acp{SNN} without impairing their performance.

\paragraph{Deep \acp{SNN} with skip connections are more robust to initialization.}
In deep \acp{ANN}, skip connections are standard practice to facilitate optimization and improve training performance \mbox{\citep{Srivastava2015-yw, Srivastava2015-xy, He2016-mi}}. 
In \acp{ResNet}, direct (skip) connections are implemented between non-successive layers.
We speculated that similar skip connections could rescue the impaired spike propagation in deep \acp{CSNN} and therefore increase their robustness to initialization.
We tested this idea in three-layer \acp{CSNN} with skip connections between each hidden layer and the readout layer (Supplementary Fig.~\ref{sfig:skipcon}a; Methods). 
Skip connections indeed increased robustness to initialization, with respect to both large $\sigma_U > 10$ and small $\sigma_U \ll 1$ (Supplementary Fig.~\ref{sfig:skipcon}b).
However, generalization performance after training did not increase as a result of added skip connections (Supplementary Fig.~\ref{sfig:skipcon}d). 
Notably, for initializations with small $\sigma_U \ll 1$, optimized networks only propagated activity through the skip connection between the first hidden layer and the readout layer, effectively reducing the network to a single hidden layer.
As skip connections did not prevent all layers from being quiescent in deep \acp{SNN}, we wondered whether homeostatic plasticity and skip connections complement each other and further increase performance for initializations with $\sigma_U\ll 1$. 
Thus, we trained three-layer \acp{CSNN} with skip connections and ongoing homeostatic plasticity.  
Networks with combined skip connections and homeostatic plasticity also exhibited an enhanced robustness to initialization, but did not show a significantly better generalization performance (Supplementary Fig.~\ref{sfig:skipcon}c, d). 
We concluded that skip connections are a viable approach to increase the robustness towards initializations with large $\sigma_U > 10$ in deep \acp{CSNN}, but are not able to compensate for vanishing gradients in deep layers when $\sigma_U \ll 1$. 

\begin{SCfigure}[10][t]
    \centering
    \includegraphics{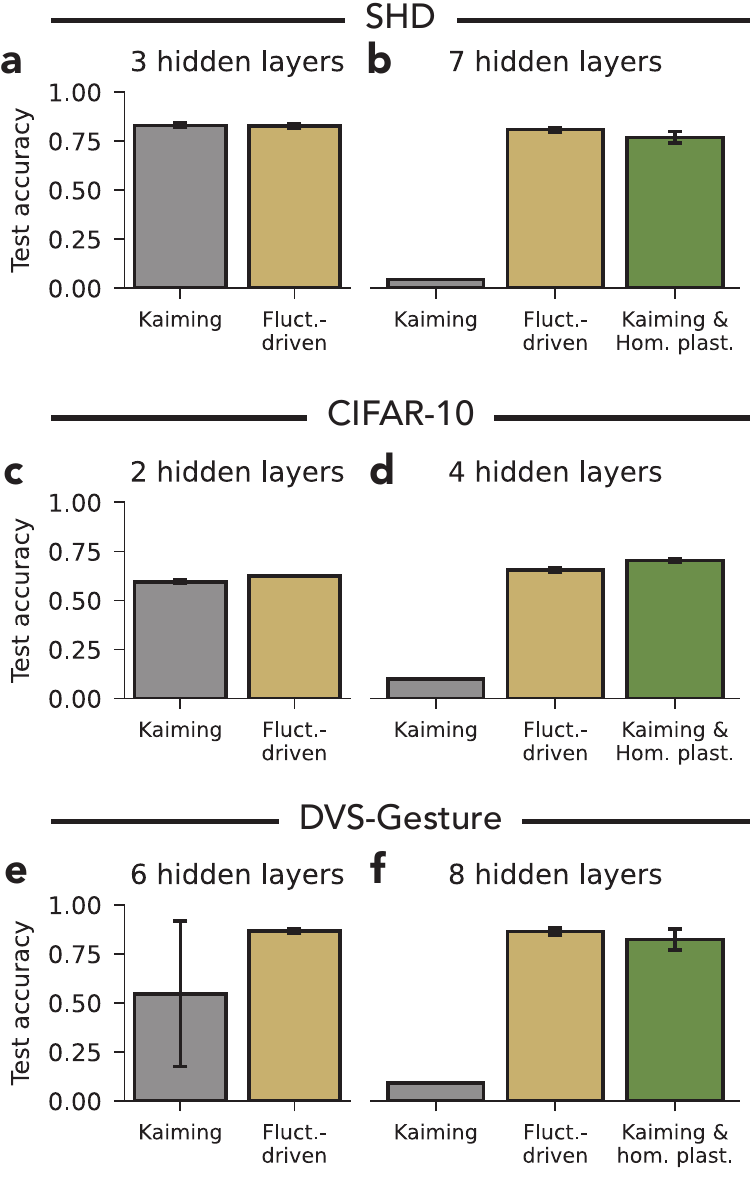}
    \caption{\textbf{Fluctuation-driven initialization enables training of deep \acp{SNN} across multiple datasets.} 
    \textbf{(a)}~Test accuracy of three-layer \acp{CSNN} trained on the SHD dataset. Networks were initialized either with standard Kaiming initialization (Kaiming) or fluctuation-driven initialization with $\sigma_U=1$. All error bars indicate standard deviation across five runs.
    \textbf{(b)}~Test accuracy of seven-layer \acp{CSNN} trained on the SHD dataset. Networks with Kaiming initialization were additionally trained with ongoing homeostatic plasticity (Kaiming \& Hom.\ plast.) 
    \textbf{(c)}~Same as panel (a), but for two-layer networks trained on the CIFAR-10 dataset.
    \textbf{(d)}~Same as panel (b), but for four-layer networks trained on the CIFAR-10 dataset.
    \textbf{(e)}~Same as panel (a), but for six-layer networks  trained on the DVS-Gestures dataset.
    \textbf{(f)}~Same as panel (b), but for eight-layer networks trained on the DVS-Gestures dataset.
    }
    \label{fig:modelcomp}
\end{SCfigure}

\paragraph{Fluctuation-driven initialization performs robustly across datasets.}
Together, our results suggest that traditional Kaiming initialization used for \acp{ANN} is sufficient for training three-layer \acp{CSNN}, but breaks down when training seven-layer or deeper \acp{CSNN} on the SHD dataset.
In contrast, our proposed initialization strategy with the target fluctuation parameter set to $\sigma_U=1$ yields close-to-optimal training performance in both three- and seven-layer networks.
To directly compare fluctuation-driven and Kaiming initialization, we measured generalization performance in terms of test accuracy after training. 
As expected, we found only small differences in test accuracy for three-layer networks (Fig.~\ref{fig:modelcomp}a; Tab.~\ref{tab:acc_deep}).

Specifically, Kaiming initialized three-layer networks achieved an average test accuracy of $83.1\%\pm1.2$ (validation accuracy: $ 95.9\%\pm 1.6$), while the same networks initialized with our proposed strategy reached an average test accuracy of $82.7\%\pm1.1$ (validation accuracy: $94.1 \%\pm1.7 $).
Seven-layer networks initialized with Kaiming initialization performed close to chance level after training (test accuracy: $4.5\%\pm 0.0$; validation accuracy: $4.7 \%\pm 0.5$), while networks initialized with $\sigma_U=1$ reached $80.9\%\pm1.2$ accuracy on the test set (Fig.~\ref{fig:modelcomp}b; Tab.~\ref{tab:acc_deep}; validation accuracy: $96.1 \%\pm 0.8$). 
As homeostatic plasticity was able to compensate for suboptimal initializations by rescuing activity propagation in three-layer \acp{CSNN}, we wondered whether these results extend to seven-layer networks.
To this end, we trained Kaiming-initialized seven-layer \acp{CSNN} with ongoing homeostatic plasticity.
Indeed, homeostatic plasticity rescued training, but test accuracy after training (test accuracy: $77.0\%\pm 3.0$; validation accuracy: $95.0 \%\pm 1.8$) was worse compared to networks initialized with $\sigma_U=1$ that were trained without homeostatic plasticity (Fig.~\ref{fig:modelcomp}b; Tab.~\ref{tab:acc_deep}).

\begin{table}[b]
\def\arraystretch{1.4}
\setlength{\tabcolsep}{5pt}
\caption{Test accuracy in percent after training networks with different number of hidden layers and different initializations (Kaiming, Kaiming with homeostatic plasticity and fluctuation-driven initialization with $\sigma_U = 1$) on the SHD, CIFAR-10 and DVS-Gesture datasets. Errors correspond to the standard deviation.}
\centering
\begin{tabular*}{\textwidth}{@{\extracolsep{\fill}}lcccccc}
\toprule
& \multicolumn{2}{c}{SHD}
& \multicolumn{2}{c}{CIFAR-10}
& \multicolumn{2}{c}{DVS-Gesture}
 \\
 & $n_{\mathrm{H}}=3$ & $n_{\mathrm{H}}=7$ & $n_{\mathrm{H}}=2$ & $n_{\mathrm{H}}=4$ & $n_{\mathrm{H}}=6$ & $n_{\mathrm{H}}=8$\\
\midrule
Kaiming &  $83.1\pm1.2$ & $4.5\pm0.0$ & $59.5 \pm 0.8 $ & $10.0 \pm 0.0 $ & $54.6 \pm 37.1 $ & $9.1 \pm 0.0 $ \\
Kaiming \& Hom. & - & $77.0  \pm 2.9 $  & - & $70.3 \pm 0.9 $  & - & $82.3 \pm 5.3 $ \\
Fluct.-driven & $82.7 \pm 1.1 $  & $80.9 \pm 1.2 $  & $62.4 \pm 0.3 $ & $65.6 \pm 1.3 $  & $86.7 \pm 1.2 $  & $86.4 \pm 1.7 $ \\
\addlinespace
\bottomrule
\end{tabular*}
\label{tab:acc_deep}
\end{table}

\sidecaptionvpos{figure}{t}

So far, we have limited our investigation to initialization-dependence to deep \acp{CSNN} trained on the SHD dataset, which is relatively small and may thus be prone to overfitting.
To test whether our findings would generalize to other tasks, we trained deep \acp{CSNN} on two additional datasets. 
First, we considered CIFAR-10, a dataset consisting of static images.
To translate static image input into spiking, we augmented the networks with an additional layer of simulated sensory neurons into which we injected the individual image pixel values as currents. 
Both bias currents and current gain were optimized end-to-end with all other network parameters (see Methods).
We then constructed deep \acp{CSNN} with increasing numbers of hidden layers.
As before, networks were either initialized with traditional Kaiming initialization or with a target membrane potential fluctuation magnitude of $\sigma_U = 1$.
We observed that networks with up to two hidden layers showed good training performance with both initializations (Fig.~\ref{fig:modelcomp}c; Tab.~\ref{tab:acc_deep}). 
When we increased the number of hidden layers to four, networks initialized with $\sigma_U = 1$ continued to show good training performance, while networks initialized with Kaiming initialization failed to train (Fig.~\ref{fig:modelcomp}d; Tab.~\ref{tab:acc_deep}).
Training on CIFAR-10 with ongoing homeostatic plasticity was able to rescue learning in Kaiming initialized \acp{SNN} with four hidden layers.

Since CIFAR-10 is a still image dataset which lacks temporal dynamics it is less well suited for assessing \acp{SNN} performance. 
To check whether our results generalize to other commonly used \ac{SNN} datasets, we considered the DVS-Gesture dataset \citep{Amir2017-ks}, which consists of short videos that depict humans performing different hand gestures.
These videos were recorded using an event camera, yielding event-based outputs that can be used to train \acp{SNN} on classification of the performed gestures.
As before, we initialized deep recurrent \acp{CSNN} with increasing number of hidden layers using either Kaiming initialization or a target $\sigma_U = 1$ and compared their test accuracy after training.
We found that networks up to six layers could be successfully trained using either Kaiming or our proposed initialization (Fig.~\ref{fig:modelcomp}e; Tab.~\ref{tab:acc_deep}).
However, in six-layer networks, initialization with $\sigma_U = 1$ yielded more reliable training performance and higher accuracy than Kaiming initialization.
When we increased the number of hidden layers to eight, networks initialized with Kaiming initialization did not train successfully, while networks initialized with a target $\sigma_U = 1$ continued to show good learning performance (Fig.~\ref{fig:modelcomp}f; Tab.~\ref{tab:acc_deep}).
Like already observed on the SHD and CIFAR-10 datasets, training of Kaiming initialized deep networks could be rescued by adding homeostatic plasticity during training. 

Taken together, these findings paint a clear pattern of initialization dependences across datasets: 
Up to a certain number of hidden layers, which is dataset dependent, Kaiming initialization yields good training performance in \acp{SNN}. 
However, when networks become too deep, vanishing \acp{SG} prevent training in networks with Kaiming initialization. 
In contrast, our proposed initialization strategy enables learning at high performance for deeper networks when the target fluctuation magnitude is set to $\sigma_U=1$.
As a complementary data-dependent strategy, homeostatic plasticity can be used to prevent vanishing gradients and rescue learning in deep networks that were initialized in a suboptimal regime.

\subsection*{Initializing \acp{SNN} that obey Dale's law}
\begin{figure}[bt]
    \centering
    \includegraphics[width=1.0\textwidth]{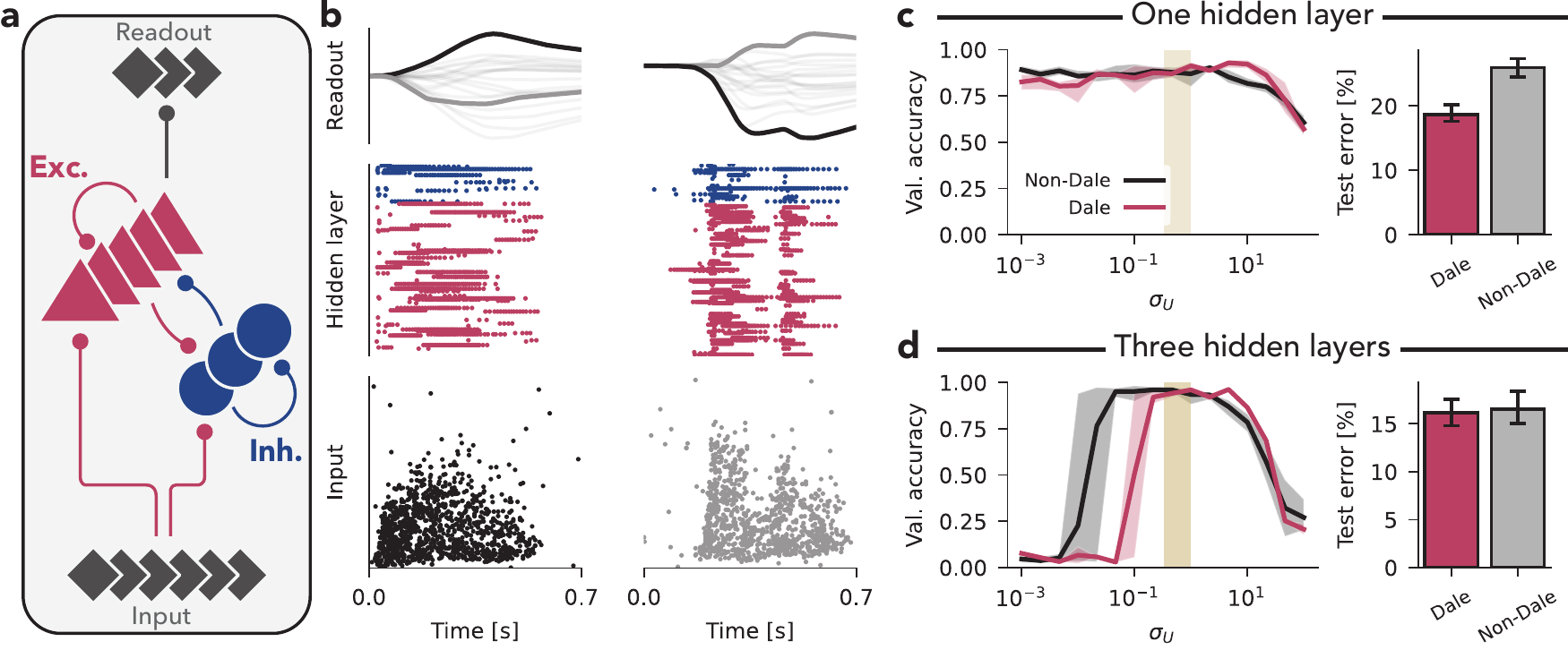}
    \caption{\textbf{Initialization of Dalian \acp{SNN} in the fluctuation-driven regime.} 
    \textbf{(a)} Schematic of a shallow \ac{SNN} obeying Dale's law. 
    Excitatory (red) and inhibitory (blue) populations are recurrently connected, but separate. 
    \textbf{(b)} Snapshot of network activity over time after training a shallow \ac{SNN} obeying Dale's law on the SHD dataset. 
    Bottom: Spike raster of input layer activity from two samples corresponding to two different classes.
    Middle: Spike raster of excitatory (red) and inhibitory (blue) activity in the hidden layer.
    Top: Membrane potential of readout units. The readout units corresponding to the two input classes are highlighted in different shades.
    \textbf{(c)} Performance comparison of Dalian and Non-Dalian shallow \acp{SNN}. 
    Left: Validation accuracy after training on the SHD dataset as a function of initialization target $\sigma_U$. The shaded region around the lines indicates the range of values across five random seeds. The sand-colored shaded region corresponds to our suggested target fluctuation magnitude $ 1\le \xi \le 3$.
    Right: Test error of the five best performing models in terms of validation accuracy, for Dalian and Non-Dalian \acp{SNN}. 
    Error bars mark $\pm$ one standard deviation.
    \textbf{(d)} As panel (c), for Dalian and Non-Dalian three-layer \acp{CSNN}.
    }
    \label{fig:dale}
\end{figure}

Neurons in biological \acp{SNN} are separated into excitatory and inhibitory populations, a constraint commonly known as Dale's Law \citep{Eccles1954-td}. 
With added biological constraints, functional \acp{SNN} constitute an important in-silico model system for computational neuroscience.
To advance the development of biologically constrained \acp{SNN}, we extended our initialization theory to \acp{SNN} obeying Dale's law (see Methods), i.e., in which each hidden layer consists of recurrently connected but separate excitatory and inhibitory populations (Fig.~\ref{fig:dale}a). 
At initialization, we require a balance between excitatory and inhibitory currents ($\mu_U = 0$), as is commonly observed in biology \cite{Rupprecht2018-sl,Spiegel2014-ra}.
To accomplish such balance, we assume that excitatory and inhibitory synaptic weights are drawn from independent exponential distributions, whose mean values are set according to our theory to ensure the desired membrane potential dynamics (Supplementary Fig.~\ref{sfig:dale-theory}).
This strategy allowed us to initialize Dalian networks with the same target $\sigma_U$ as non-Dalian networks.

To test whether Dalian networks in the fluctuation-driven regime could be trained to high accuracy like their non-Dalian counterparts, we first considered fully connected recurrent Dalian \acp{SNN} with one hidden layer trained on the SHD dataset.
Dalian networks initialized with $\sigma_U=1$ accurately solved the SHD task after training for 200 epochs ($99.8\%\pm0.0$ train \& $82.2\%\pm1.2$ test accuracy; Fig.~\ref{fig:dale}b). 
Next, to test the robustness to initialization in Dalian networks, we initialized Dalian \acp{SNN} with different targets $\sigma_U$ and trained them on the SHD dataset.
For direct comparison between Dalian \acp{SNN} and non-Dalian \acp{SNN}, we constructed \acp{SNN} with a total of $n_\mathrm{h}=160$ hidden layer neurons, which were further split into $n_\mathrm{exc}=128$ and $n_\mathrm{inh}=32$ neurons for the Dalian case.
After training, the Dalian networks exhibited similar robustness to initialization as non-Dalian networks (Fig.~\ref{fig:dale}c).
While we did not observe a large difference between Dalian and non-Dalian networks in validation accuracy, Dalian networks exhibited higher accuracy on the SHD test dataset.
This result suggests that the separation into excitatory and inhibitory populations could provide a functionally beneficial constraint for \acp{SNN} with one recurrently connected hidden layer trained on the SHD dataset.

We wondered whether the better generalization performance of shallow Dalian \acp{SNN} would extend to deeper \ac{CSNN} network architectures. 
To address this question, we constructed Dalian \acp{CSNN} with three hidden layers. 
Again, networks were initialized with different targets $\sigma_U$ and trained on the SHD dataset.
We found that Dalian \acp{CSNN} with three hidden layers were more sensitive to initialization than their non-Dalian counterparts (Fig.~\ref{fig:dale}d).
However, when successfully trained, Dalian and Non-Dalian \acp{CSNN} resulted in similar test accuracies. 

In summary, our initialization strategy extends to Dalian \acp{SNN} with different network architectures and enables robust training on the SHD dataset.
Unexpectedly, constraining networks with Dale's law increased generalization accuracy by 7.1\% in shallow networks. 
However, this effect did not generalize to deep \acp{CSNN}. 
Thus initializing Dalian networks in the fluctuation-driven regime is beneficial for their training and it will be interesting future work to study whether and how these findings generalize to larger datasets.

\section*{Discussion}

We have introduced a general and easy-to-implement initialization strategy for \acp{SNN} and shown that it yields close-to-optimal performance across different \ac{SNN} architectures and datasets. 
To that end, we developed a simple and general theory based on the notion of fluctuation-driven firing and tested it empirically in numerical simulations.
We found that shallow \ac{SNN} architectures are surprisingly robust to initialization with small synaptic weight magnitudes, whereas deep \acp{CSNN} require carefully chosen initial weight distributions that our theory accurately predicts. 
Further, our analysis showed that suboptimal initial weight choices result in vanishing or exploding \acp{SG}, similar to \acp{ANN}.
Importantly, for all network architectures, including deep convolutional, recurrent, and Dalian \acp{SNN}, and the different datasets we considered, we found that fluctuation-driven initialization with given target membrane fluctuations of $\sigma_U=1$, resulted in stable activity propagation and close-to-optimal learning performance. 
Based on our results, we recommend to initialize \acp{SNN} in the fluctuation-driven regime using a target $\sigma_U=1$ for all practical purposes.
If activity propagation remains limited after training, a problem we observed in deeper network architectures, we recommend the addition of firing rate homeostasis to the training loss either for the entire training process or transiently during an in initial priming period.

\medskip

Functional \acp{SNN} are most commonly obtained by converting a previously trained ANN \citep{Esser2015-mi,Hunsberger2016-em,Cao2015-yt,OConnor2013-uo,Bu2022-ab} or through direct training using timing-based methods \citep{Bohte2002,booij_gradient_2005, Mostafa2018,Kheradpisheh2020, Comsa2020} or \acp{SG} \citep{Zenke2018-id,Zenke2021-zg,Neftci2019-ie}. 
While both approaches can result in well-performing networks, direct training typically leads to sparser activity levels while also leveraging spike timing which can be beneficial for energy efficiency \citep{davidson_comparison_2021}. 
The initialization strategy developed in this article mainly applies to direct training approaches and specifically for \acp{SNN} trained with \acp{SG}. 

Most previous \ac{SNN} studies relied on weight initialization strategies that were established for \acp{ANN} in which they aim at keeping the variance of gradients constant through time or layers. 
For example, Xavier (Glorot) initialization \citep{Glorot2010-dj} achieves stable variance in the backward pass by appropriately scaling the initial weight distribution. 
While the Xavier initialization was originally developed for linear networks, the Kaiming (He) initialization \citep{He2015-kv} extends this approach by explicitly taking into account the \ac{ReLU} non-linearity, thereby enabling the training of deeper \ac{ReLU} networks. 
While both Xavier and Kaiming initialization posit a scaling of weights by the number of input neurons as $~\sim \nicefrac{1}{n_\mathrm{in}}$, their profound effects on learning performance largely result from the different choice of absolute weight scale, which differs by a factor of two, a direct consequence of neuronal non-linearity.

Alternatively, the weight scale in the case of \acp{SNN} is often determined empirically, however, there are some proposed initialization strategies, although they often lack a sound theoretical foundation.
For example, \citet{Lee2016} proposed to normalize the magnitude of back-propagated errors across layers by initializing synaptic weights from a uniform distribution $W_{(l)} \sim \mathcal{U} [ -\sqrt{\nicefrac{3}{n_l}}, \sqrt{\nicefrac{3}{n_l}}]$, where $n_{l}$ is the number of incoming synapses.
However, this approach was only validated in networks with two hidden layers trained on an event-based version of the MNIST dataset \citep{Orchard2015-rl} and requires manual tuning of a per-layer weight scale to define the spiking threshold.
\citet{Bellec2018-yu} in turn initialized weights for spiking LSTM models from a Normal distribution as $W_{(l)} \sim \mathcal{N} \left( 0, \nicefrac{1}{n_{l-1}} \right)$, whereas \citet{Zenke2021-zg} used a uniform distribution $W_{(l)} \sim \mathcal{U} [ -\sqrt{\nicefrac{1}{n_{l-1}}}, \sqrt{\nicefrac{1}{n_{l-1}}}]$. 
A more intricate approach was developed by \citet{Herranz-Celotti2022}, who suggested a number of conditions on the initial weights that aim, e. g., to balance the variance of the gradients across time and layers. 
Based on those conditions, the authors derived a way to determine the weight scale for a Uniform distribution. 
While initialization with an ad-hoc chosen weight scale can support successful training in shallow networks, none of these studies applied their initialization strategies to network architectures with more than two hidden layers.
However, as shown in  this article, shallow network architectures are intrinsically robust to initialization as long as the weights are small enough while the need for \ac{SNN}-specific initialization mainly arises when training deep \acp{SNN}.
It thus remains an open question whether these results generalize to deep \acp{SNN}.

Recently, \citet{Ding2021-ql} proposed an initialization strategy that generalized to deep \acp{SNN} architectures.
The authors related the magnitude of backpropagating gradients in feed-forward \acp{SNN} to the synaptic weight distribution and proposed a weight scale for normally distributed synaptic weights that takes into account some parameters of neuronal dynamics, but does not consider dataset-dependent input parameters. 
While similar to the approach outlined here, this initialization strategy is limited to centered weight distributions and feed-forward networks.
In addition, this particular study limited the forward pass to 20 time steps and delta synapses, compared to 100-500 time steps and current-based synapses in our simulations.
Using delta synapses and a smaller number of time steps can increase the performance of \acp{SNN}, but does so at the cost of biologically realistic membrane potential dynamics.
How well these results generalize to recurrently connected \acp{SNN} or more biologically plausible membrane dynamics is unclear.

Our fluctuation-driven initialization strategy follows a similar approach to \citet{Glorot2010-dj, He2015-kv} by setting a target variance for neuronal activity.
However, due to the non-continuous nature of the spiking nonlinearity, we formulated the goal in terms of the membrane potential variance $\sigma_U$ instead of the post-nonlinearity activation.
Our theory results in weight scaling that not only accounts for the number of hidden layer neurons, but also data- and architecture-dependent parameters.

In contrast to the above approaches, \citet{Mishkin2015-zd} proposed an iterative initialization strategy to achieve unit variance of neuronal activations at each layer during a pre-training period.
The implementation of a pre-training period is similar to the homeostatic priming period we applied here.
However, instead of setting an explicit target for the population variance, our homeostatic regularizer targeted per-neuron spiking activity with the goal to enable activity propagation.

\medskip
Our work has several limitations.
First, our theory is limited to \ac{LIF} neurons with current-based synapses. 
Although the current-based \ac{LIF} is by far the most commonly used neuron model in \acp{SNN}, its synaptic dynamics can allow for biologically implausible and undesirable membrane potential values. 
Indeed, we found that some neurons exhibit exceptionally small ($u(t) \ll 0$) or large ($u(t) \gg \theta$) membrane potential values after training, which were not intended when designing the \acp{SG}.
Future work could explore the possibility of using conductance-based synapses or additional regularization losses to constrain membrane potentials to a biologically plausible range while still allowing for large simulation time steps and thus rapid training.

Second, we performed numerical simulations with a relatively large time step of $\Delta t = 2$\,ms. 
Choosing the simulation time step marks a trade-off between computational efficiency on one side and sensitivity of the membrane potential to quickly changing inputs on the other. 
Indeed, better performing deep \acp{SNN} have been trained using a time step on the order of the membrane potential time constant \citep{Na2022-vb,Ding2021-ql}. 
Our choice of simulation time step reflects a compromise between minimizing computation time and allowing for sufficiently realistic membrane dynamics. 

Third, our initialization theory for recurrent \acp{SNN} and \acp{SNN} following Dale's law, rests on the assumption of balanced input currents, i.e., $\mu_U = 0$, similar to what is observed in neurobiology \citep{brunel_dynamics_2000, vogels_neural_2005}.
Whether and how this balanced state contributes to initial learning phases in the brain remains an open question for experimental and theoretical neuroscience.
However, in our numerical simulations, sweeps across the parameters of initial weight distribution in shallow \acp{SNN} (Fig.~\ref{fig:shallow}) suggest that a slight dominance of inhibition over exciation may represent a similarly favorable or even more advantageous initial state for learning.
Therefore, it equally remains to be clarified whether unbalanced currents, for example by a slight dominance of inhibition at initialization, could further support learning in functional \acp{SNN} models. 

Fourth, our fluctuation-driven initialization theory makes several assumptions that could be violated in some use cases using real-world data. 
Our theory assumed that all input neurons are independent from each other and fire according to a homogeneous Poisson process with a common firing rate $\nu$.
Although we have shown that the systematic bias from violating this assumption in the Randman and \ac{SHD} datasets is not too large (Fig.~\ref{fig:valid}), other datasets with a different spatiotemporal structure could lead to destructive deviations from the theory.
As a result, the current initialization strategy could be improved by taking into account more complex firing statistics of the input data.
Additionally, our derivations neglected the spike reset of \acp{LIF} neurons.
While mathematically more complex, it would be possible to consider the reset dynamics in our derivations using a Fokker-Plank approach \citep{Amit1997-ds}.
However, given that the deviations from the theory due to spatio-temporal structure in the data likely outweigh the contribution of the spike reset, it is questionable whether this extension would confer an advantage.

Finally, we assumed equal firing rates $\nu = \nu_{\mathrm{dataset}}$ for all neurons in a layer in deep \acp{SNN}, and for both excitatory and inhibitory populations in Dalian \acp{SNN}.
Despite being violated for most initialization targets (cf.\ Fig~\ref{fig:deep}), this simplification allowed for effective initialization with a common target $\sigma_U=1$ across multiple datasets with vastly different average firing rates (see Methods).
Interestingly, for initialization with target $\sigma_U=1$ on the SHD dataset, we indeed observed relatively constant firing rates across layers.
A consistent method to estimate the firing rate distribution in deep layers at the time of initialization could improve the performance of other initialization targets and could potentially enable training of deeper \acp{SNN}.
As an alternative approach, dynamic initialization during a pre-training priming period could be extended to adjust weights by regularizing the output firing rate to a target value in an iterative fashion.
Similar to approaches that have been proposed for \acp{ANN} \citep{Mishkin2015-zd}, such an iterative and dynamic initialization strategy could enable activity propagation and learning in even deeper \acp{SNN}.
However, increasing the number of layers in recurrently connected \acp{SNN} did not lead to significant performance improvements in our study. 
Given the success of deep \acp{ANN}, this suggests either that the datasets used to evaluate \acp{SNN} are too simple, or that deep \acp{SNN} architectures and learning algorithms are still in their infancy and could be significantly improved.

\medskip

In conclusion, the fluctuation-driven initialization proposed in this article 
facilitates training of diverse \acp{SNN} architectures in neuromorphic engineering and computational neuroscience by striking a balance between seamless applicability and learning performance.
Our work also adds further support to the idea that the fluctuation-driven firing regime, which is widely observed in the brain, may serve as an optimal initial state for future learning, and specifically for scenarios in which learning can be seen as an end-to-end optimization problem \citep{marblestone_toward_2016, richards_deep_2019, lillicrap_backpropagation_2020}. 
While our work only provides the first step toward more effective \ac{SNN} initialization, it opens up several future exciting directions such as initialization in the presence of sparse connectivity, neuronal cell-type diversity, and suggests that we should take a deeper look at the role of homeostatic plasticity in dynamically preparing networks for optimal learning performance.

\section*{Methods}

\subsection*{Learning tasks}

We trained \acp{SNN} on several synthetic and real-world
classification problems with increasing computational complexity and from different input modalities (auditory, visual) to test our initialization strategy. The properties of the different datasets are summarized in Supplementary Tab.~\ref{stab:datasets}.

\paragraph{Synthetic random manifolds (Randman).} 
We used a versatile synthetic classification dataset based on precise input spike timings drawn from smooth 
random manifolds as previously described (\citealp{Zenke2021-zg};  \href{https://github.com/fzenke/randman}{https://github.com/fzenke/randman}). 
The approach allows for flexible dataset generation with different degrees of complexity by varying the number of classes, the intrinsic manifold dimension $D$, the smoothness parameter $\alpha$, and the embedding space dimension $M$. 

Here, we chose parameters to ensure the problem could not be solved by an \ac{ANN} without a hidden layer. 
Specifically, we set the embedding space dimension $M = n_{\mathrm{randman}} = 20$, \mbox{$D=\alpha=1$} and generated spike trains of 100\,ms duration with $10$ different classes for all our simulation experiments. 
To account for delays in activity propagation through the network, we appended 100\,ms of no spiking activity to the generated inputs, resulting in a total duration of $T_{\mathrm{randman}}=200$\,ms and hence an average input firing rate of $\nu_{\mathrm{randman}}=5$\,Hz. Further, we used the same random seed to generate the dataset for all experiments in which we compare different initializations to avoid variability due to differences in the dataset. 
Specifically, we generated a 10-way classification dataset with 1000 samples for each class, 800 of which served as training data and two sets of 100 samples each served as validation and testing data, respectively.

\paragraph{Spiking Heidelberg Digits (SHD).} 
The SHD dataset \citep{cramer_heidelberg_2020} is a real-world auditory dataset containing recordings of spoken digits $(0-9)$ in both German and English from different speakers.
It is freely available for download at \href{https://zenkelab.org/datasets}{https://zenkelab.org/datasets}. 
To obtain input spikes, the raw audio data was pre-processed by a biologically inspired cochlear model \citep{cramer_heidelberg_2020} and mapped into an $n_{\mathrm{SHD}}=700$ dimensional input space. 
As individual input samples are of different duration, we considered only the first $T_{\text{SHD}}=700$\,ms of each sample, which corresponds to a \mbox{fraction $>98$ \%} of all input spikes. 
Spliced inputs were binned into $\frac{T_{\text{SHD}}}{\Delta t}$ time steps and fed directly into the \acp{SNN}. 
We used a random subset corresponding to 10\% of the training data as a validation set.
To evaluate generalization performance, we finally used the standard SHD test dataset which contains data from separate speakers that were not included in the training dataset.

\paragraph{CIFAR-10.} 
The CIFAR-10 dataset consists of 3x32x32 pixel images belonging to 10~different classes (6000 images for each class) and is commonly used as a visual classification dataset for neural networks \citep{Krizhevsky2009-ru}. 
The first dimension of the input data corresponds to the three RGB color channels.
As an image dataset it does not have an intrinsic time dimension in the input.
To translate static images in temporal spiking input, we designed an additional sensory neuron encoding layer, placed in between the input layer and the first hidden layer, that converts static images into spike trains. 
First, each input pixel data was repeated with a fan-out factor of five along the channel dimension, thereby creating an effective input dimension of 15x32x32 for each image.
Second, each pixel value was multiplied by a gain factor to which a bias term was added before using the result as a current input to an encoding layer consisting of 15x32x32 \ac{LIF} units. 
The encoding weights for height and width dimensions were tied across all encoding units, leading to an encoding weight matrix of shape 15x1x1.
Thus, each encoding neuron receives the weighted pixel value of a single color channel as a constant synaptic current, and transduces this value into an output spike train.
Synaptic weights of the encoding layer were not subject to our initialization strategy, but randomly drawn from a normal distribution with mean $0$ and standard deviation $\frac{\Delta t}{\tau_{\text{syn}} \sqrt{n_{l-1}}}$.  
Biases of encoding units were randomly drawn from a normal distribution with mean $0$ and standard deviation $\frac{1}{\sqrt{n_{\text{enc}}}}$ and optimized end-to-end during training.
For training, CIFAR-10 images were transformed to a normalized range $[-1, 1]$ and presented as input to the encoding layer for a duration of $T_{\text{CIFAR-10}}=100$\,ms. 
To obtain an estimate of the firing rate required for the initialization of hidden layers, we measured the average population firing rate of the encoding layer in response to the CIFAR-10 training dataset at the time of initialization, resulting in $\nu_{\text{CIFAR-10}} = 14.3$\,Hz.

\paragraph{DVS128 Gesture Dataset.} 
The DVS-gesture dataset \citep{Amir2017-ks} is a standard benchmark for event-based processing. 
It consists of 1342 videos of 11 different hand and/or arm gestures that were recorded with a biologically inspired Dynamic Vision Sensor (DVS), yielding sparse and asynchronous input spike trains. 
The data from 23 recorded subjects serve as training data, while the data from 6 separate subjects serve as test data. 
Before training, we applied data augmentation and down-sampling, more specifically (1) random omission of events, (2) down-sampling of the original recordings and (3) random temporal crop. 
First, recorded (binary) events were dropped with a probability of $p=0.5$. 
Second, the original 2x128x128 pixels recordings were down-sampled to 2x32x32 pixels.
Third, a random 1-second fragment was extracted from each sample. These 1-second long segments were then binned into $\frac{1000\,\text{ms}}{\Delta t}$ time steps and used as input to the SNN for $T_{\text{DVS-Gestures}}=1000\,\text{ms}$.

\subsection*{Network models}

All \ac{SNN} models were trained with \acp{SG} using PyTorch \citep{NEURIPS2019_9015}. To
this end, we used custom software written in Python~3.6.9 (\href{https://github.com/fmi-basel/stork}{https://github.com/fmi-basel/stork}). 
For numerical simulations, all models were implemented in discrete time with time step $\Delta t = 2$\,ms.
This time step was a compromise between numerical integration accuracy and computational and memory efficiency during training.

\paragraph{Neuron model.}
All units were implemented as simple \ac{LIF} neurons with exponential current-based synapses \citep{Gerstner2014-ke}. 
In discrete time, the membrane potential of neuron~$i$ in layer~$l$ is characterized by the update equation
\begin{equation}
	U^{(l)}_i [n+1] = \left(\lambda_{\text{mem}} U^{(l)}_i[n] + \left(
1-\lambda_{\text{mem}} \right) I^{(l)}_i[n] \right) \left( 1 - S^{(l)}_i[n]
\right) ~,
\label{eq:lif-update}
\end{equation}
where $U^{(l)}_i[n]$ is this neuron's membrane potential at time step $n$ and $S^{(l)}_i[n]$ is the associated binary (spiking) output of this neuron defined as $S^{(l)}_i[n] = \Theta \left(U^{(l)}_i[n] -\theta \right)$ with spike threshold $\theta$, where $\Theta$ is the Heaviside step function. 
For simplicity, we set $\theta=1$, so that the resting membrane potential is zero and the firing threshold is equal to one. 
The membrane decay variable $\lambda_{\text{mem}}$ is determined by the membrane time constant $\tau_{\text{mem}}$ through
$\lambda_{\text{mem}} \equiv \exp \left(-\frac{\Delta t}{\tau_{\text{mem}}}\right)$. 
Lastly, $I^{(l)}_i[n]$ denotes the incoming synaptic current to neuron $i$ at time step $n$ and is defined as
\begin{equation}
	I^{(l)}_i[n+1] = \lambda_{\text{syn}} I^{(l)}_i[n] + \sum_j
w_{ij}^{(l)}S^{(l-1)}_j[n] + \sum_j v_{ij}^{(l)}S^{(l)}_j[n]
\end{equation}
with the feed-forward weight matrix $W$ and optional recurrent weight matrix $V$. 
The synaptic decay variable $\lambda_{\text{syn}}$ is related to the synaptic time constant through $\lambda_{\text{syn}} \equiv \exp \left(-\frac{\Delta t}{\tau_{\text{syn}}}\right)$. 
The neuronal parameters used throughout our simulations can be found in Tab.~\ref{tab:neuron-params}.

\begin{table}[htpb]
\def\arraystretch{1.4}
\setlength{\tabcolsep}{5pt}
\caption{Neuronal parameters $\tau_{\text{mem}}$ and $\tau_{\text{syn}}$ used in the numerical simulations of \acp{SNN}, where in the case of Dalian \acp{SNN} we provide the values for the time constants of the exc./inh. populations.}
\centering
\begin{tabular*}{0.66\textwidth}{@{\extracolsep{\fill}}lcc}
\toprule
                            & Non-Dalian \acp{SNN}      & \makecell{Dalian \acp{SNN}\\ (exc. / inh.)}\\
 \midrule
 $\tau_\text{mem}$ [ms]			& 20				    & 20 / 10	\\
 $\tau_\text{syn}$ [ms]		    & 10			    	& 10 / 5 	\\
 \bottomrule
 \end{tabular*}
\label{tab:neuron-params}
\end{table}

\paragraph{Readout units.} 
The units in the readout layer are identical to the above neuron model, but were not allowed to spike. 
Additionally, the membrane time constant of readout units $\tau_{\text{out}}$ could be different from the hidden layer units. 
Unless otherwise mentioned, we set $\tau_{out}= T_{\text{data}}$ for all simulations to allow readout units to integrate inputs over the entire stimulus duration. 

\paragraph{Dale's Law.}
In \acp{SNN} obeying Dale's law (cf. Fig.~\ref{fig:dale}), each hidden layer consists of independent excitatory ($E$) and inhibitory ($I$) populations of \ac{LIF} neurons with membrane time constants $\tau_{\text{mem}}^{\text{E}}$ and $\tau_{\text{mem}}^{\text{I}}$, respectively. 
In discrete time, the membrane potential of each excitatory or inhibitory neuron $i$ in layer $l$ is identical to Eq. ~\eqref{eq:lif-update}, where $\lambda_\text{mem}$ is replaced with the population-specific decay variables $\lambda_\text{mem}^E$ and $\lambda_\text{mem}^I$, respectively (Tab.~\ref{tab:neuron-params}).
Like in the non-Dalian case, the decay variables are related to the membrane time constants as $\lambda^{E}_{\text{mem}} \equiv \exp \left(-\frac{\Delta t}{\tau^E_{\text{mem}}}\right)$ and $\lambda^{I}_{\text{mem}} \equiv \exp \left(-\frac{\Delta t}{\tau^I{\text{mem}}}\right)$.
In contrary to the non-Dalian case, the input currents in Dalian \acp{SNN} consist of separate excitatory and inhibitory components originating from distinct presynaptic populations.
For both excitatory and inhibitory populations, the input current can therefore be decomposed into feed-forward excitatory ($F$), recurrent excitatory ($R$) and recurrent inhibitory ($I$) components, such that
\begin{eqnarray}
I^{(l), E}_i[n] &=& I^{(l), FE}_i[n] + I^{(l), RE}_i[n] - I^{(l), IE}_i[n] \\
I^{(l), I}_i[n] &=& I^{(l), FI}_i[n] + I^{(l), RI}_i[n] - I^{(l), II}_i[n] ~.
\end{eqnarray}
Thus, both the excitatory and inhibitory populations receive two sources of excitatory and one source of inhibitory input. In discrete time, the incoming synaptic currents to the excitatory neuron $i$ of layer $l$ are given as
\begin{eqnarray}
	 I^{(l), FE}_i[n+1] &=& \lambda^{E}_{\text{syn}} I^{(l), FE}_i[n] + \sum_j w_{ij}^{(l), FE}S^{(l-1), E}_j[n] \\
	 I^{(l), RE}_i[n+1] &=& \lambda^{E}_{\text{syn}} I^{(l), RE}_i[n] + \sum_j w_{ij}^{(l), RE}S^{(l), E}_j[n] \\
	 I^{(l), IE}_i[n+1] &=& \lambda^{I}_{\text{syn}} I^{(l), IE}_i[n] + \sum_j w_{ij}^{(l), IE}S^{(l), I}_j[n] ~,
\end{eqnarray}
where $\lambda^{E}_{\text{syn}}$ and $\lambda^{I}_{\text{syn}}$ are the decay variables of excitatory and inhibitory currents, which are related to their respective synaptic time constants $\tau^E_{\text{syn}}$ and $\tau^I_{\text{syn}}$ (Tab.~\ref{tab:neuron-params}) as described before. 
Similarly, the synaptic currents into inhibitory neuron $i$ of layer $l$ are defined as
\begin{eqnarray}
	 I^{(l), FI}_i[n+1] &=& \lambda^{E}_{\text{syn}} I^{(l), FI}_i[n] + \sum_j w_{ij}^{(l), FI}S^{(l-1), E}_j[n] \\
	 I^{(l), RI}_i[n+1] &=& \lambda^{E}_{\text{syn}} I^{(l), RI}_i[n] + \sum_j w_{ij}^{(l), RI}S^{(l), E}_j[n] \\
	 I^{(l), II}_i[n+1] &=& \lambda^{I}_{\text{syn}} I^{(l), II}_i[n] + \sum_j w_{ij}^{(l), II}S^{(l), I}_j[n] ~.
\end{eqnarray}
Together, the dynamics of each Dalian hidden layer are therefore determined by two feed-forward weight matrices $W^{FE}$ and $W^{FI}$ and four recurrent weight matrices $W^{RE}$, $W^{RI}$, $W^{IE}$ and $W^{II}$.

\paragraph{Connectivity.} 
Feed-forward and recurrent networks were all-to-all connected without bias terms unless mentioned otherwise.  
We used two types of  convolutional networks with 1-dimensional and 2-dimensional convolutional kernels (Supplementary Tab.~\ref{stab:architecture_CSNN}). 
Recurrent connections in \acp{CSNN} were implemented as convolutions with filter kernels of size five and a stride of one. For weight initialization of \acp{CSNN}, we set $n=\text{fan}_\text{in}$, the number of inputs to each filter. 

In \acp{SNN} and \acp{CSNN} obeying Dale's law, weights between consecutive layers and recurrent weights within hidden layers were constrained to be positive both at initialization and continuously during training, with the exception of readout weights, which were not sign constrained.
Both excitatory and inhibitory populations in hidden layers received feed-forward inputs from the excitatory population of the previous layer. 
All networks obeying Dale's law were fully recurrent, featuring recurrent connections within and between excitatory and inhibitory populations in each layer (E$\rightarrow$E, E$\rightarrow$I, I$\rightarrow$I and I$\rightarrow$E). 

\paragraph{Skip connections.}
We implemented skip connections as additional all-to-all connections between each except the last hidden layer and the readout layer, such that the readout units receive a separate input from every hidden layer (cf. Supplementary Fig.~\ref{sfig:skipcon}). 

\paragraph{Supervised loss function.} 
All networks were trained by minimizing a standard cross-entropy loss
\begin{equation}
\mathcal{L}_{\text{sup}} = - \frac{1}{K} \sum_{k=1}^K \sum_{c=1}^C y_c^k~,
\log\left(p_c^k\right)
\end{equation}
where the one-hot encoded target for input $k$ is denoted by $y_c^k$, $K$ is the number of input samples and $C$ is the number of classes. 
The associated output probabilities $p_c^k$ are given by the Softmax function
\begin{equation}
	p_c^k = \frac{\exp({a_c^k)}}{\sum_{i=1}^C \exp(a_i^k)} ~.
\end{equation}
The scores $a_c^k$ for each input $k$ are dependent on the membrane potential of the associated readout units $U_c^{\text{(out)}}$ and can take different forms. 
For all simulations in this paper, we use the maximum value over all time steps $a_c^k = \max_n\left(U_c^{\text{(out)}}[n]\right)$.

\paragraph{Activity regularization.}
Unless otherwise mentioned, all networks were subject to activity regularization to constrain spiking activity. 
To that end, we added loss terms corresponding to a soft upper bound on the population-level spiking activity for each layer $l$ as
\begin{equation}
	g^{(l),k}_{\text{upper}} = \left( \left[\frac{1}{M^{(l)}} \sum_i^{M^{(l)}}
\zeta_i^{(l),k} - v_{\text{upper}} \right]_+ \right)^2~,
\end{equation}
where $\zeta_i^{(l), k} = \left( \sum_n^N S_i^{(l), k}[n]\right)$ is the spike count of neuron $i$ in layer $l$ given input sample $k$ and $M^{(l)}$ is the number of neurons in hidden-layer $l$.
In 1-dimensional \acp{CSNN} receiving auditory inputs, we set $M^{(l)} = n_{\text{features}}^{(l)} \times n_{\text{neurons}}^{(l)}$ and, similarly, in 2-dimensional \acp{CSNN} receiving visual inputs, we set $M^{(l)} = n_{\text{features}}^{(l)} \times n_{\text{x}}^{(l)} \times n_{\text{y}}^{(l)}$ with  $n_{\text{x}}^{(l)}$ and $n_{\text{y}}^{(l)}$ denoting the number of $x$ and $y$ coordinates in the layer, respectively.
This activity regularization effectively prevents the population-level activity from exceeding the threshold spike count $v_{\text{upper}}$, which we set to $v_{\text{upper}} = \frac{T_{\text{data}}}{100}$ to achieve an upper bound average population firing rate of 10\,Hz per layer.
The regularization loss in case of population-level upper bound for spiking activity $\mathcal{L}_{\text{UB}}$ would thus be
\begin{equation}
\mathcal{L}_{\text{UB}} = -\lambda_{\text{upper}} \sum_l^L
g^{(l),k}_{\text{upper}}~,
\end{equation}
where $\lambda_{\text{upper}}$ denotes the strength of the regularization.

\paragraph{Homeostatic plasticity.} In networks with homeostatic
plasticity (cf.\ Figs.~\ref{fig:homplast} and \ref{fig:modelcomp}) we added an additional term
to the total loss acting as a per-neuron
lower bound on the spiking activity. This per-neuron lower bound loss on
spiking activity $\mathcal{L}_{\text{HP}}$ was defined as
\begin{eqnarray}
g^{(l),k}_{\text{lower}} &=& \frac{1}{M^{(l)}} \sum_i^{M^{(l)}}\left(
\left[-\left(\zeta_i^{(l),k} - v_{\text{lower}}\right) \right]_+ \right)^2\\
\mathcal{L}_{\text{HP}} &=& -\lambda_{\text{lower}} \sum_l^L
g^{(l),k}_{\text{lower}}~,
\end{eqnarray}
where the first equation describes the per-neuron loss term for each layer $l$.
$\zeta_i^{(l),k}$ corresponds to the spike count of neuron $i$ in layer $l$,
$M^{(l)}$ is the number of neurons in layer $l$, $v_{\mathrm{lower}}$ denotes a
lower bound on the spike count and $\lambda_{\mathrm{lower}}$ is the
regularizer strength. \\
With $v_{\mathrm{lower}}=1$, this additional regularization term penalizes
neurons that do not spike and thus ensures spiking activity in each neuron. By setting $v_{\mathrm{lower}}$ to other
positive values one may achieve a desired lower bound on the per-neuron firing rate.

\paragraph{Surrogate gradient descent.} 
To minimize the loss $\mathcal{L}$, we adjusted network parameters in the direction of the negative \ac{SG}. 
We computed \acp{SG} for the parameter updates using \ac{BPTT} and the automatic differentiation capabilities of PyTorch \citep{NEURIPS2019_9015}. 
Because the spiking non-linearity of the spiking neuron model is not differentiable, we approximate its derivative \begin{equation}
S^{\prime}\left(U^{(l)}_i[n]\right) = \Theta' \left(U^{(l)}_i[n] - \theta \right)
\end{equation}
with the surrogate \begin{equation}
\tilde{S}^{\prime}\left(U^{(l)}_i[n]\right) = h\left(U^{(l)}_i[n] - \theta\right)~. 
\end{equation}
Throughout this study, we use the SuperSpike surrogate non-linearity \citep{Zenke2018-id}
\begin{equation}
	h(x) =\frac{1}{\left(\beta |x|+1\right)^2}
\end{equation}
with steepness parameter $\beta = 20$.
For the simulations of deep \acp{CSNN} with a rescaled \ac{SG} nonlinearity reported in Supplementary Fig.~\ref{sfig:surrgrad-rescaled}, we used the re-scaled surrogate derivative 

\begin{equation}
\tilde{S}^{\prime}\left(U^{(l)}_i[n]\right) = \frac{h\left(U^{(l)}_i[n] - \theta\right)}{h\left(\theta\right)}~.
\end{equation}
In this case, the surrogate derivative at rest is equal to one, i.e., $\tilde{S}^{\prime}\left(0 \right) = 1$, where "at rest" refers to the absence of input to the corresponding neuron, causing its membrane potential to remain at zero. Thus, using this rescaled nonlinearity, and in the absence of any membrane potential fluctuations, gradient magnitudes do not decay during backpropagation over the inactive layers.

\paragraph{Optimizer.}
We used the SMORMS3 optimizer \cite{Funk2015-xl} unless mentioned otherwise. Given a parameter $\theta$, SMORMS3 performs the following update step after every mini batch:
\begin{eqnarray}
	g_1^{(\theta)} &:=& \left(1 - r^{(\theta)} \right) g_1^{(\theta)} + r^{(\theta)} \left(\frac{\partial \mathcal{L}}{\partial \theta}\right) \\
	g_2^{(\theta)} &:=& \left(1 - r^{(\theta)} \right) g_2^{(\theta)} + r^{(\theta)} \left( \frac{\partial \mathcal{L}}{\partial \theta} \right)^2 \\
	m^{(\theta)} &:=& 1 + m^{(\theta)} \left(1 - \frac{\left(g_1^{(\theta)}\right)^2}{g_2^{(\theta)} + \epsilon} \right)	~,
\end{eqnarray}
where $r^{(\theta)} = \frac{1}{m^{(\theta)}+1}$ and $\epsilon = 1 \times 10^{-16}$ is a small positive value to avoid division by zero. 
Before the first training epoch, the optimizer state variables are initialized as $g_1^{(\theta)} = g_2^{(\theta)} = 0$ and $m^{(\theta)} = 1$. 
The parameter update after each mini batch is then performed as
\begin{eqnarray}
	\Delta \theta = - \left(\frac{\partial \mathcal{L}}{\partial \theta}\right) \min \left( \eta, \frac{\left(g_1^{(\theta)}\right)^2}{g_2^{(\theta)} + \epsilon} \right) \frac{1}{\sqrt{g_2^{(\theta)}} + \epsilon}~,
\end{eqnarray}
where $\eta$ is the base learning rate.

\subsection*{Weight initialization}

For fluctuation-driven initialization of synaptic weight parameters, the \ac{PSP}-kernel parameters $\bar\epsilon$ and $\hat\epsilon$ introduced in Equations \eqref{eq:main-mu-u} and \eqref{eq:main-sigma-u} can be computed analytically or numerically (see Supplementary Material \ref{sup:kernel}). 
Because we used a relatively large time step of $\Delta t = 2$\,ms 
for which there are non-negligible differences between the two, we used the numerical integration values for all simulations as they are closer to the actual simulation (Tab.~\ref{tab:epsilon-numerical}).

For strictly feed-forward networks, the fluctuation-driven initialization strategy was already covered in the main text. 
In the following, we derive the extensions to deep convolutional \acp{SNN}, recurrent \acp{SNN} and \acp{SNN} obeying Dale's law.

\begin{table}[htpb]
\def\arraystretch{1.4}
\setlength{\tabcolsep}{5pt}
\caption{Values of the \ac{PSP}-kernel integrals $\bar\epsilon$ and $\hat\epsilon$ used for weight initialization in the numerical simulations, rounded to four decimal places. Due to the large simulation time step of $\Delta t = 2$\,ms, $\bar\epsilon$ and $\hat\epsilon$ were obtained numerically. The analytical expressions for $\bar\epsilon$ and $\hat\epsilon$ can be found in Supplementary Tab.~\ref{stab:epsilon}.}
\centering
\begin{tabular*}{0.66\textwidth}{@{\extracolsep{\fill}}lcc}
\toprule
                            & Non-Dalian \acp{SNN}      & \makecell{Dalian \acp{SNN}\\ (exc. / inh.)}\\
 \midrule
 $\bar\epsilon$			    & 0.0110				    & 0.0110 / 0.0061	\\
 $\hat\epsilon$ 		    & 0.0020			    	& 0.0020 / 0.0012 	\\
 \bottomrule
 \end{tabular*}
\label{tab:epsilon-numerical}
\end{table}

  \paragraph{Fluctuation-driven initialization of recurrent networks.} For the
initialization of recurrent layers, we introduce the additional parameter $0 <
\alpha < 1$, that determines the proportion of membrane potential fluctuations
caused by \emph{feed-forward} connections in contrast to \emph{recurrent}
connections:
\begin{equation}
    \alpha = \frac{\text{Part of }\sigma_U^2 \text{ caused by feed-forward connections}}{\text{Total }\sigma_U^2}~.
    \label{eq:alpha_nd}
\end{equation}
To this end, we consider a postsynaptic LIF neuron receiving
feed-forward input from $n_F$ neurons with firing rate $\nu_F$ and recurrent
input from $n_R$ hidden layer neurons with average firing rate $\nu_R$.
Feed-forward weights are initialized as $W \sim \mathcal{N}(\mu_W, \sigma_W^2)$
and recurrent weights are initialized as $V \sim \mathcal{N}(\mu_V,
\sigma_V^2)$ The mean $\mu_U$ and variance $\sigma_U^2$ of the membrane
potential are then given by
\begin{eqnarray}
\mu_U &=& n_F \mu_W \nu_{F}\bar\epsilon + n_R \mu_V \nu_{R}\bar\epsilon \\
\sigma^2_{U}&=&n_F(\sigma_{W}^2 + \mu_W^2)\nu_F\hat\epsilon + n_R(\sigma_{V}^2+\mu_V^2)\nu_R\hat\epsilon~.
\end{eqnarray}
In practice the firing rate $\nu_R$ of the hidden layer is
difficult to predict due to finite size effects. 
Hence, we make the simplifying assumption $\nu = \nu_F
= \nu_R = \nu_{\text{dataset}}$. In other words, we assume that the average
firing rate of the hidden layers is equal to the input firing rate.

Since we want $\alpha$ to control the membrane potential \textit{fluctuations}
only, which are determined by $\sigma_W^2$ and $\sigma_V^2$, we can initialize
recurrent and feed-forward weights with a common mean, i.e., $\mu_{WV}$ =
$\mu_W$ = $\mu_V$
defined as
\begin{equation}
\label{eq:muWV_nondale_rec}
\mu_{WV} = \frac{\mu_U}{\left(n_F + n_R\right) \nu_{\text{dataset}}
\bar\epsilon}
\end{equation}
and subsequently solve for $\sigma_W^2$ and $\sigma_V^2$ independently:
\begin{eqnarray}
\label{eq:sigmaW_nondale_rec}
\sigma^2_W &=& \frac{\alpha}{n_F \nu \hat\epsilon} \left(\frac{\theta -
\mu_U}{\xi} \right)^2 - \mu_{WV}^2
\\	
\label{eq:sigmaV_nondale_rec}
\sigma^2_V &=& \frac{1-\alpha}{n_R \nu \hat\epsilon} \left(\frac{\theta -
\mu_U}{\xi} \right)^2 - \mu_{WV}^2~.
\end{eqnarray}
In this article, we used $\alpha=0.9$ for all simulations with recurrently connected
hidden layers unless stated otherwise, so that the majority of membrane potential fluctuations originate from
feed-forward input.
  
\paragraph{Fluctuation-driven initialization of Dalian networks.} Networks
following Dale's law consist of separate excitatory and inhibitory populations whose output weights are sign constrained.
To initialize the sign constrained connections, we relied on exponential or log-normal weight distributions instead of normally distributed weights, where the choice of a log-normal distribution is inspired by
findings from neurobiology \citep{buzsaki_log-dynamic_2014}.

Parameterizing the excitatory and inhibitory weight distributions with $\lambda$ for the exponential and $\mu$ for the log-normal distribution, respectively, allows us to obtain explicit expressions for the initial weight distributions leading to the target membrane potential fluctuations with mean $\mu_U$ and variance $\sigma_U^2$. 
Unless stated otherwise, the weights in Dalian \acp{SNN} were initialized using the exponential distribution throughout the numerical simulations. While we provide here the expression for weight initialization using the exponential distribution, a derivation for log-normally distributed initial weights can be found in the Supplementary Material \ref{sec:lognormal}.

We start by observing that, regardless of the weight distribution from which synaptic weights are sampled, mean and variance of the membrane potential of a neuron $i$ in a Dalian network are
defined as
\begin{eqnarray}
\mu_U^{(i)} &=& \sum_j^{n_E} w^{E}_{ij} \nu_E \bar\epsilon_E -
\sum_k^{n_I}
w^{I}_{ik} \nu_I \bar\epsilon_I
\\
\left(\sigma^{(i)}_U\right)^2 &=& \sum_j^{n_E} (w^{E}_{ij})^2 \nu_E
\hat{\epsilon}_{E} +
\sum_k^{n_I} (w^{I}_{ik})^2 \nu_I \hat{\epsilon}_{I} ~,
\end{eqnarray}
where we assume equal firing rates $\nu_E$ and $\nu_I$ for all excitatory and
inhibitory neurons in our experiments, respectively.

For weights drawn from exponential distributions, i.e., $w^E \sim
\text{Exp}(\lambda_E)$ and $w^I \sim \text{Exp}(\lambda_I)$ with mean
$\frac{1}{\lambda}$ and variance $\frac{1}{\lambda^2}$, we can rewrite mean and variance of the membrane potential for each neuron as
\begin{eqnarray}
\mu_U &=& \frac{n_E\nu_E\bar\epsilon_E}{\lambda_E} -
\frac{n_I\nu_I\bar\epsilon_I}{\lambda_I}
\\
\sigma_U^2 &=& n_E
\left(\frac{1}{\lambda_E^2}+\left(\frac{1}{\lambda_E}\right)^2\right) \nu_E
\hat{\epsilon}_{E} +
n_I \left(\frac{1}{\lambda_I^2}+\left(\frac{1}{\lambda_I}\right)^2\right) \nu_I
\hat{\epsilon}_{I}
\nonumber
\\
&=& \frac{2 n_E\nu_E\hat\epsilon_E}{\lambda_E^2} + \frac{2
n_I\nu_I\hat\epsilon_I}{\lambda_I^2}  ~.
\label{eq:exp-sigma_u}
\end{eqnarray}
We further assume that the target $\mu_U = 0$, as would be expected in balanced
networks.
From the definition of $\mu_U$, we obtain an explicit
relationship between $\lambda_I$ and $\lambda_E$
\begin{equation} \label{eq:lambda_inh}
	\lambda_I = \lambda_E \frac{n_I \nu_I\bar\epsilon_I}{n_E\nu_E \bar\epsilon_E}~,
\end{equation}
which we use to define a combined E/I ratio based on network parameters
\begin{equation}
	\Delta_{EI} = \frac{n_I \nu_I \bar\epsilon_I}{n_E \nu_E \bar\epsilon_E}~.
	\label{eq:nd_delta_ei}
\end{equation}
Substitution of this relationship into equation \eqref{eq:exp-sigma_u} gives
us 
\begin{equation}
\sigma^2_U=\frac{2 n_E\nu_E\hat\epsilon_E}{\lambda_E^2} + \frac{2
n_I\nu_I\hat\epsilon_I}{(\lambda_E \Delta_{EI})^2}  ~.
\end{equation}
Finally we can solve the above for $\lambda_E$
\begin{equation}\label{eq:lambda_exc}
	\lambda_E = \frac{\sqrt{2(\Delta_{EI}^2 n_E\nu_E\hat\epsilon_E +
n_I\nu_I\hat\epsilon_I)}}{\sigma_U\Delta_{EI}} ~.
\end{equation}
Together, equations \eqref{eq:lambda_inh} and \eqref{eq:lambda_exc} allow us to
parameterize excitatory and inhibitory weights as a function of $\sigma_U$, taking
into account data- and network-dependent parameters, which is summarized in Tab.~\ref{tab:nondalian_init}.
Note that this initialization relies on a target membrane potential mean
$\mu_U = 0$.

\begin{table}[htpb]
\def\arraystretch{1.4}\setlength{\tabcolsep}{5pt}
\caption{Summary of strategies for fluctuation-driven initialization of \acp{SNN}.}
\centering
\begin{tabular*}{\textwidth}{@{\extracolsep{\fill}}lclc}
\toprule
\makecell{\textbf{Network} \\ \textbf{Architecture}}
& \makecell{\textbf{Weight} \\ \textbf{Distribution}} 	
& \makecell{\textbf{Weight} \\ \textbf{Parameters}} 	
& \makecell{\textbf{Good regime for} \\ \textbf{Initialization}} 
 \\
\midrule
\addlinespace
\multirow{3}{*}{Feed-forward} 

& \makecell{Centered: \\ $W\sim \mathcal{N}\left(0, \sigma_W^2\right)$}
& $\begin{aligned} 
            \sigma^2_W = \frac{\sigma_U^2}{n \nu \hat\epsilon}
    \end{aligned}$
& $\begin{aligned}
\frac{1}{3}\le \sigma_U& \le 1
    \end{aligned}$
\\
\addlinespace
\cmidrule(l){2-4}
& \makecell{Non-centered: \\ $ W\sim \mathcal{N}\left(\mu_W,
\sigma_W^2\right)$}
& $\begin{aligned}
           \mu_W &= \frac{\mu_U}{n \nu \bar\epsilon}  \\
\sigma^2_W &= \frac{1}{n \nu \hat\epsilon} \left(\frac{\theta -
\mu_U}{\xi}\right)^2 - \mu_W^2
    \end{aligned}$ 
& $\begin{aligned}
    \mu_U&<\theta\\ 
    1\le\xi&\le 3
    \end{aligned}$
\\
\addlinespace
\midrule
\addlinespace
\multirow{5}{*}{Recurrent} 

& \makecell{Centered: \\
            $\begin{aligned}
                W &\sim \mathcal{N}\left(0, \sigma_W^2\right) \\
                V &\sim \mathcal{N}\left(0, \sigma_V^2\right)
            \end{aligned}$} 
& $\begin{aligned}
           \sigma^2_W &= \alpha \frac{\sigma_U^2}{n_F \nu \hat\epsilon} \\
\sigma^2_V &= \left(1 - \alpha\right) \frac{\sigma_U^2}{n_R \nu \hat\epsilon}
\\
    \end{aligned}$ 
& $\begin{aligned}
\frac{1}{3}\le\sigma_U&\le 1\\ 
        0<\alpha&<1
    \end{aligned}$
\\
\addlinespace
\cmidrule(l){2-4}
& \makecell{Non-centered: \\
            $\begin{aligned}
                W &\sim \mathcal{N}\left(\mu_{WV}, \sigma_W^2\right) \\
                V &\sim \mathcal{N}\left(\mu_{WV}, \sigma_V^2\right)
            \end{aligned}$}   
& $\begin{aligned}
           \mu_{WV} &= \frac{\mu_U}{\left(n_F+n_R\right) \nu \bar\epsilon}  \\
\sigma^2_W &= \frac{\alpha}{n_F \nu \hat\epsilon} \left(\frac{\theta -
                \mu_U}{\xi} \right)^2 - \mu_{WV}^2 \\
\sigma^2_V &= \frac{1 - \alpha}{n_R \nu \hat\epsilon} \left(\frac{\theta -
                \mu_U}{\xi} \right)^2 - \mu_{WV}^2
    \end{aligned}$ 
& $\begin{aligned}
        \mu_U&<\theta\\
        1\le\xi&\le 3\\
        0<\alpha&<1
    \end{aligned}$\\
\addlinespace
\bottomrule
\end{tabular*}
\label{tab:nondalian_init}
\end{table}

\paragraph{Fluctuation-driven initialization of Dalian networks with excitatory recurrence.}
Dalian layers are always recurrently connected, as they require a connection between the separate excitatory and inhibitory populations in each layer. For this reason, the Dalian network from the above paragraph has recurrent inhibitory connections ($I \rightarrow E$ and $I \rightarrow I$). Here, we consider the case of additional excitatory recurrence, i.e $E \rightarrow I$ and $E \rightarrow E$ connections. 

Again, we require inhibitory currents to balance excitatory currents on average to achieve a mean membrane potential $\mu_U = 0$. 
Additionally, similar to non-Dalian \acp{SNN} with recurrent connections, the parameter $\alpha$ describes the proportion of excitatory membrane potential fluctuations that are caused by feed-forward excitatory connections, whereas the proportion of recurrent excitation is given by $(1-\alpha)$.
For the derivation, we consider a single neuron in a Dalian layer, receiving one recurrent inhibitory ($I$), one feed-forward excitatory ($F$) and one recurrent excitatory ($R$) input connection. In this setting, mean $\mu_U$ and variance $\sigma_U^2$ of the membrane potential of that neuron are given by

\begin{eqnarray}
\mu_U &=& \frac{N_{F}\nu_F\bar\epsilon_{E}}{\lambda_F} + \frac{N_{R}\nu_R\bar\epsilon_{E}}{\lambda_R} - \frac{N_{I}\nu_I\bar\epsilon_{I}}{\lambda_I} 
\label{eq:dr_mu_u}\\
\sigma^2_{U}&=&\frac{2 N_{F}\nu_F\hat\epsilon_{E}}{\lambda_F^2} + \frac{2 N_{R}\nu_R\hat\epsilon_{E}}{\lambda_R^2} + \frac{2 N_{I}\nu_I\hat\epsilon_{I}}{\lambda_I^2} ~ .
\label{eq:dr_sigma_U}
\end{eqnarray}
For the sake of simpler notation, we assume $\nu = \nu_F=\nu_R=\nu_I$ in this derivation. 
We also made this assumption of equal firing rates in the application of this initialization strategy in our numerical simulations. Since it is not possible to estimate the firing rates of excitatory and inhibitory hidden neuron populations in advance, we chose $\nu= \nu_{\text{dataset}}$.

The ratio of membrane potential fluctuations caused by the excitatory feed-forward connections compared to the total excitatory input, which we defined as $\alpha$, can explicitly be written as
\begin{equation}
\alpha = \frac{\text{Part of }\sigma_U^2 \text{ caused by excitatory feed-forward connections}}{\text{Part of }\sigma_U^2 \text{ caused by all excitatory connections}} = \frac{ \frac{2 N_{F}\nu\hat\epsilon_{E}}{\lambda_F^2} }{ \frac{2 N_{F}\nu\hat\epsilon_{E}}{\lambda_F^2} + \frac{2 N_{R}\nu\hat\epsilon_{E}}{\lambda_R^2} } ~ ,
\end{equation}
which we can solve for $\lambda_R$ to obtain
\begin{equation}
\lambda_R = \lambda_F \sqrt{ \frac{\alpha N_R}{N_F - \alpha N_F} } = \lambda_F\Delta_{R}  ~,
\label{eq:dr_lambda_r}
\end{equation}
where we introduced the scalar $\Delta_R$ to make subsequent notation easier. We can then insert Eq.~\eqref{eq:dr_lambda_r} into Eq.~\eqref{eq:dr_mu_u}
\begin{equation}
\mu_U = \frac{N_{F}\nu\bar\epsilon_{E}}{\lambda_F} + \frac{N_{R}\nu\bar\epsilon_{E}}{\lambda_F\Delta_{R}} - \frac{N_{I}\nu\bar\epsilon_{I}}{\lambda_I}
\end{equation}
to receive an expression for $\lambda_I$

\begin{equation}
    \lambda_I = \lambda_F \frac{ \Delta_R \bar\epsilon_I N_I }{\Delta_R \bar\epsilon_E N_F + \bar\epsilon_E N_R} = \lambda_F\Delta_{EI}^R  ~.
    \label{eq:dr_lambda_i}
\end{equation}
We introduce here again a network-parameter dependent scalar $\Delta_{EI}^R$. Using the scalars  $\Delta_R$ and $\Delta_{EI}^R$, we can now substitute both $\lambda_I$ and $\lambda_R$ in Eq.~\eqref{eq:dr_sigma_U} to obtain

\begin{equation}
\sigma^2_{U} = \frac{2 N_{F}\nu\hat\epsilon_{E}}{\lambda_F^2} + \frac{2 N_{R}\nu\hat\epsilon_{E}}{\left(\lambda_F\Delta_R\right)^2} + \frac{2 N_{I}\nu\hat\epsilon_{I}}{\left(\lambda_F\Delta_{EI}^R\right)^2}  ~,
\end{equation}
which can be solved for $\lambda_F$
\begin{equation}
    \lambda_F = \frac{ \sqrt{2 \nu \left((\Delta_{EI}^{R})^2 \hat\epsilon_E N_R + \Delta_R^2\left(\Delta_{EI}^R N_F \hat\epsilon_E + N_I \hat\epsilon_I \right) \right)} }{ \sigma_U \Delta_{R} \Delta_{EI}^R}~.
    \label{eq:dr_lambda_f}
\end{equation}

Equations~\eqref{eq:dr_lambda_r},~\eqref{eq:dr_lambda_i} and~\eqref{eq:dr_lambda_f} let us parameterize the initial feed-forward excitatory (F), recurrent excitatory (R) and recurrent inhibitory (I) weight distributions as a function of the target membrane potential fluctuations $\sigma_U$ with a mean membrane potential $\mu_U = 0$. The suggested weight distributions including their parameters and a range of values for a good initialization are summarized in Supplementary Tab.~\ref{stab:dalian_init}. 

\paragraph{Kaiming (He) initialization.}
We implemented Kaiming (He) initialization as described by \citet{He2015-kv}. 
This commonly used strategy was originally derived for \acp{ANN} with ReLU nonlinearities and 
suggests to draw the initial weights from a centered normal distribution
\begin{equation}
     W \sim \mathcal{N}\left(0, \frac{2}{n}\right)~,
\end{equation}
where $n$ is the number of neurons and the weights have mean zero and variance
$\sigma_W^2 = \frac{2}{n}$.

\clearpage
\section*{Acknowledgments}
This work was supported by the Novartis Research Foundation. 

\section*{Author contributions}
F.Z.\ conceived the study. 
J.R., J.G.\ and F.Z.\ wrote simulation code.
J.R.\ and J.G.\ performed simulations and analyses. 
J.R., J.G., and F.Z.\ wrote the manuscript.

\section*{Competing interests}
The authors declare no competing interests.

\printbibliography

\begin{refsection}

\renewcommand{\thefigure}{S\arabic{figure}}
\setcounter{figure}{0}    

\renewcommand{\thetable}{S\arabic{table}}
\setcounter{table}{0}    

\renewcommand{\thesubsection}{S\arabic{subsection}}

\clearpage
\section*{Supplementary Figures}

\begin{figure}[htb]
	\includegraphics[width=1\textwidth]{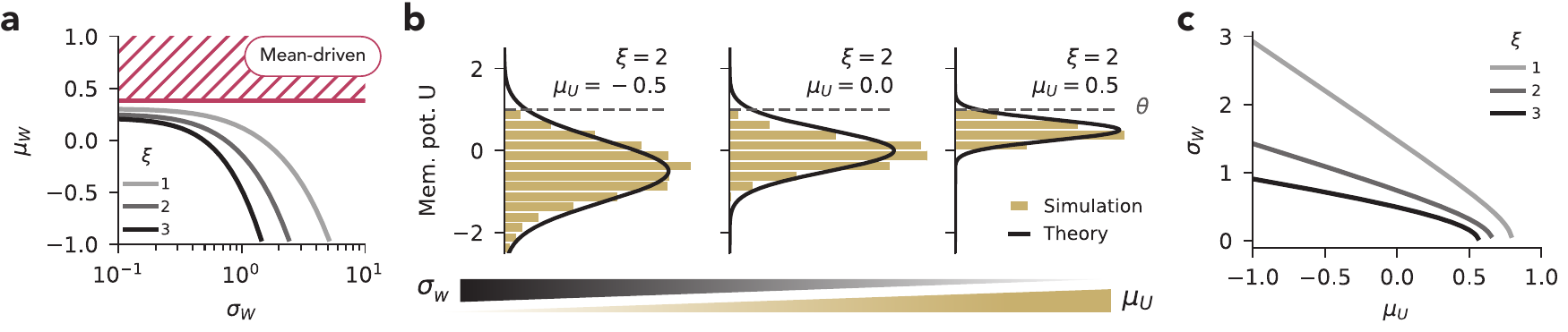}
\caption{\textbf{Initialization in the fluctuation-driven regime with non-zero $\mathbf{\mu_U}$.}
	\textbf{(a)} Different fluctuation targets $\xi$ plotted in the space spanned by the parameters of a non-centered Gaussian weight distribution $W \sim \mathcal{N}(\mu_W, \sigma_W)$. The red region indicates the regime of mean-driven initialization, where $\mu_U > \theta$. The border between fluctuation- and mean-driven regime is dependent on data and network parameters.
	\textbf{(b)} Expected and observed distributions of the membrane potential for different values of the target membrane potential mean $\mu_U$. All three displayed initializations lead to fluctuations of the same magnitude $\xi=2$.
	\textbf{(c)} The standard deviation of the weights $\sigma_W$ as a function of the target $\mu_U$ for different fluctuation targets $\xi$.
}
\label{sfig:theory-mu}
\end{figure}

\begin{figure}[htb]
	\includegraphics{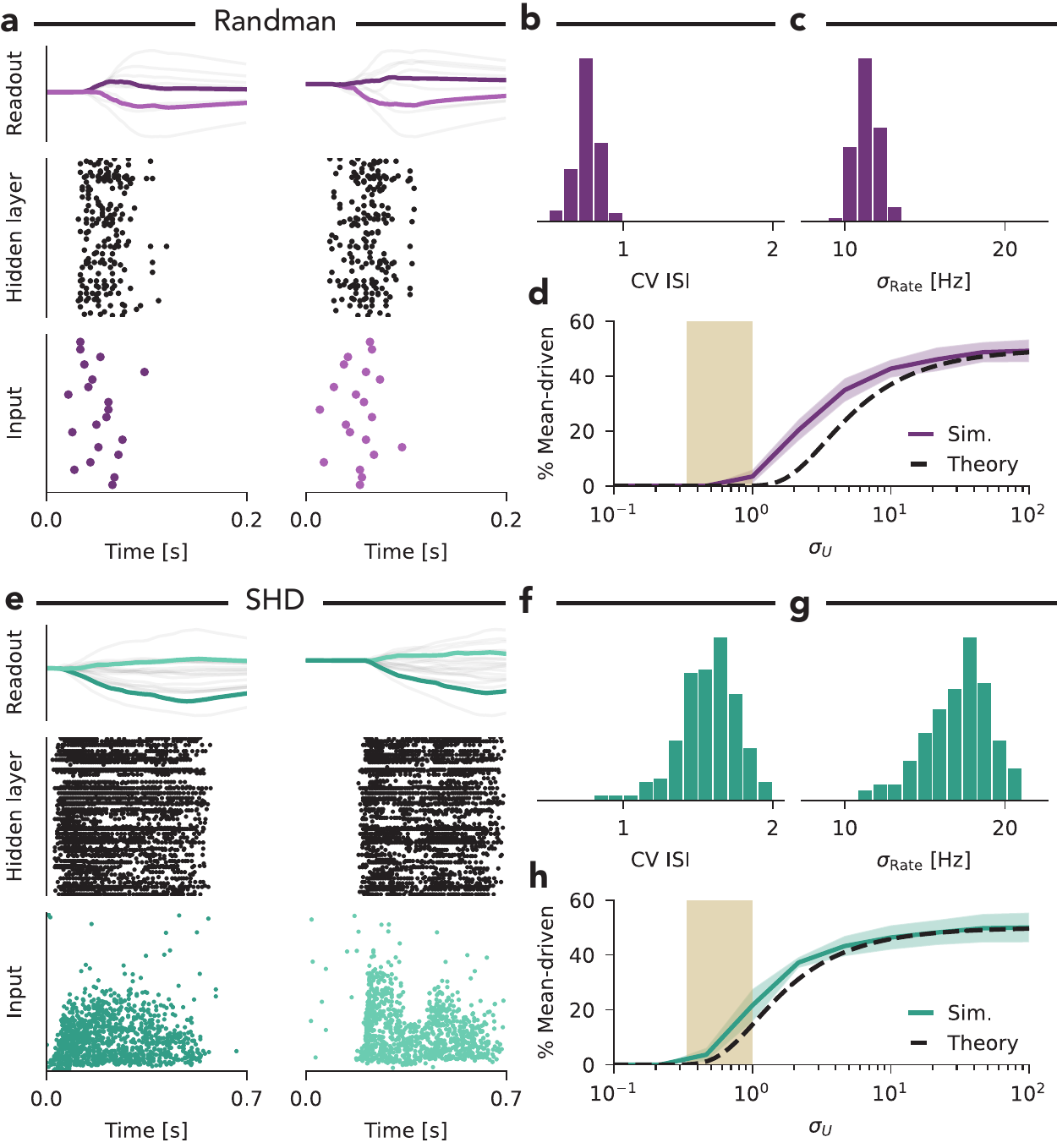}
\caption{
\textbf{Activity of shallow \acp{SNN} at time of initialization in the fluctuation-driven regime.}
	\textbf{(a)} Snapshot of activity over time before training  on the Randman dataset for an \ac{SNN} with one hidden layer. Bottom: spike raster of input layer activity from two different samples corresponding to two different classes. Middle: Spike raster of hidden layer activity. Top: Membrane potential of readout units. The readout units corresponding to the two input classes are highlighted in different shades. The network was initialized with a target $\sigma_U=1$.
	\textbf{(b)} Distribution of the coefficient of variation (CV) of inter-spike intervals (ISIs) of hidden layer neurons (see Methods).
	\textbf{(c)} Standard deviation of the population firing rate filtered with a time constant of 5\,ms (see Methods). The histogram depicts the distribution of $\sigma_{\mathrm{Rate}}$ across different input samples. 
	\textbf{(d)} Theoretically expected (Supplementary Material \ref{sup:popvar}) and numerically observed proportion of mean-driven neurons at the time of initialization as a function of the target fluctuation magnitude $\sigma_U$, for networks that are initialized to be trained on the Randman dataset ($\pm 1$ standard deviation). The sand-colored shaded region indicates the target regime $\nicefrac{1}{3} \leq \sigma_U \leq 1$.
	\textbf{(e)-(h)} Same as panels (a)-(d), for the SHD dataset.
}
\label{sfig:shallow}
\end{figure}

\begin{figure}[htb]
	\includegraphics[width=1\textwidth]{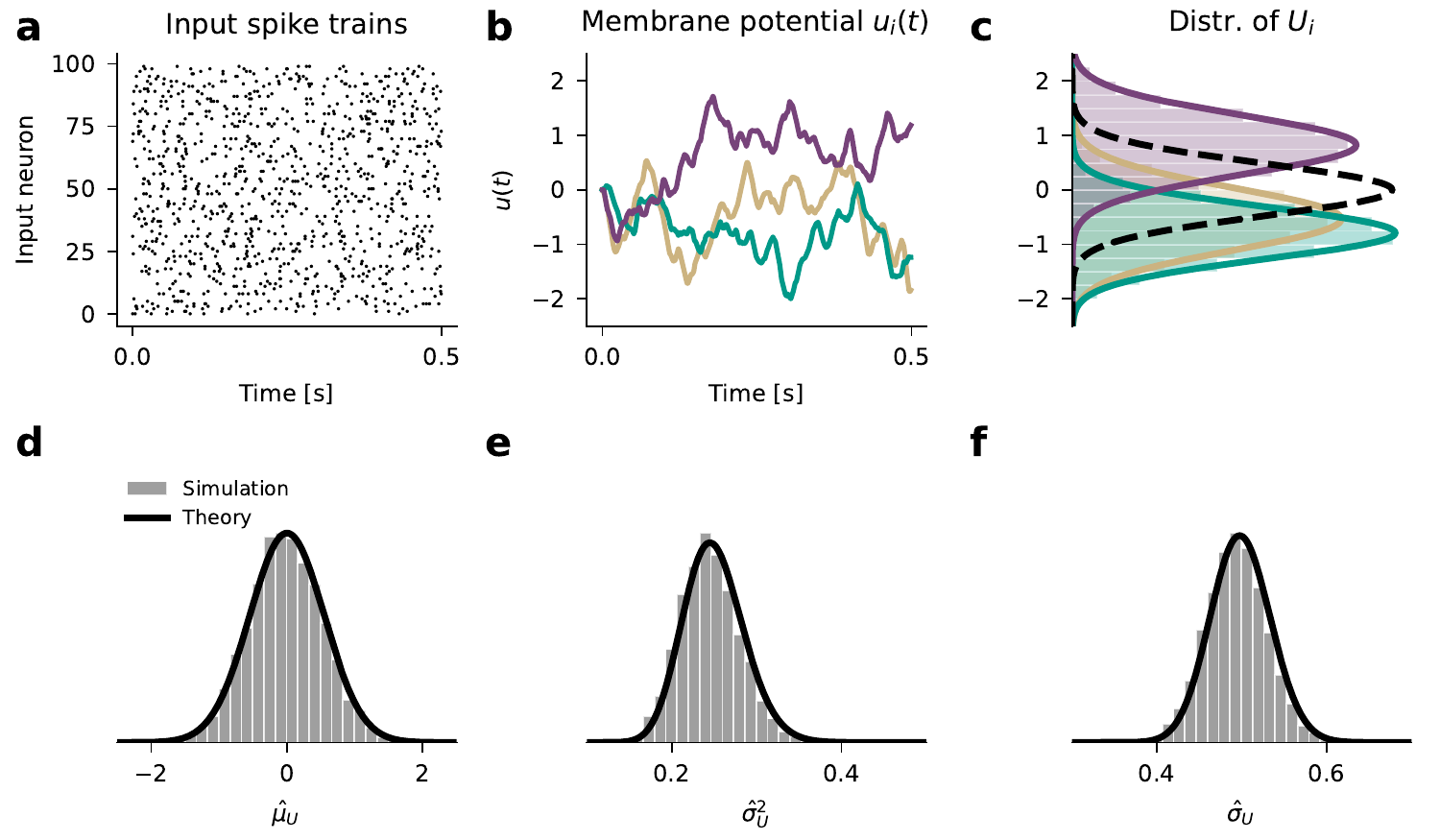}
\caption{\textbf{Population-level variability induced by random sampling of synaptic weights.}
	\textbf{(a)} Poisson input spike trains.
	\textbf{(b)} Membrane potentials $u_i(t)$ of three example neurons that were initialized with the same target $\mu_U=0$ and $\sigma_U=\nicefrac{1}{2}$.
	\textbf{(c)} Corresponding distributions of the membrane potentials $U_i$ for each of the three example neurons in panel (B). The black dashed line indicates the target membrane potential distribution $U \sim \mathcal{N}(\mu_U, \sigma_U^2)$. Note that the observed means $\hat \mu_{U_i}$ of the three
membrane potential distributions deviate from the target.
\textbf{(d)} The analytically expected and numerically observed distribution of $\hat \mu_U$ follows a Gaussian. Numerical simulations consider 5000 postsynaptic neurons initialized with the same target $\mu_U=0$ and $\sigma_U=\nicefrac{1}{2}$. 
Even when the target is set clearly in the fluctuation-driven regime, a proportion of neurons can be expected to be mean-driven.
\textbf{(e)} The observed membrane potential variances $\hat \sigma_U^2$ follow a Gamma distribution.
\textbf{(f)} The observed standard deviations of the membrane potential $\hat \sigma_U$ are Nakagami distributed. For the derivations of analytical solutions in panels (d)-(e), see Supplementary Material \ref{sup:popvar}.
}
\label{sfig:popvar}
\end{figure}

\begin{figure}[htb]
	\includegraphics{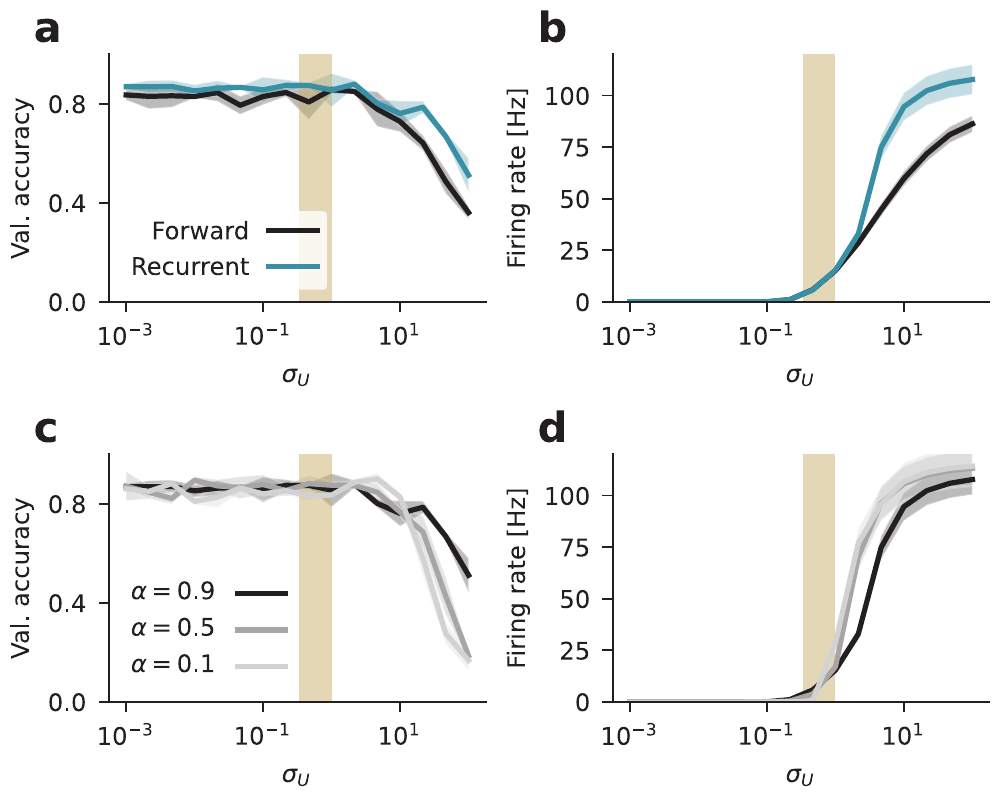}
\caption{
\textbf{Fluctuation-driven initialization of recurrent \acp{SNN}.}
	\textbf{(a)} Validation accuracy after training as a function of the target fluctuation magnitude $\sigma_U$ at initialization for \acp{SNN} with one hidden layer featuring only forward or additional recurrent connections. The shaded region around the lines indicates the range of values across five random seeds. The sand-colored shaded region corresponds to our suggested target fluctuation magnitude $ \frac{1}{3}\le \sigma_U\le 1$.
	\textbf{(b)} Population firing rate of hidden layer neurons at the time of initialization as a function of $\sigma_U$. Recurrent connections cause the firing rate to increase through a positive feedback loop when the initial fluctuation magnitude is large.
	\textbf{(c)} As panel (a), for recurrent \acp{SNN} initialized with different relative magnitudes of recurrent connections $\alpha$ (see Methods). Large values of $\alpha$ increase the contribution of feed-forward connections to membrane potential fluctuations. 
	\textbf{(d)} As panel (b), for for recurrent \acp{SNN} initialized with different values of $\alpha$.
}
\label{sfig:recurrent}
\end{figure}

\begin{figure}[htb]
	\includegraphics{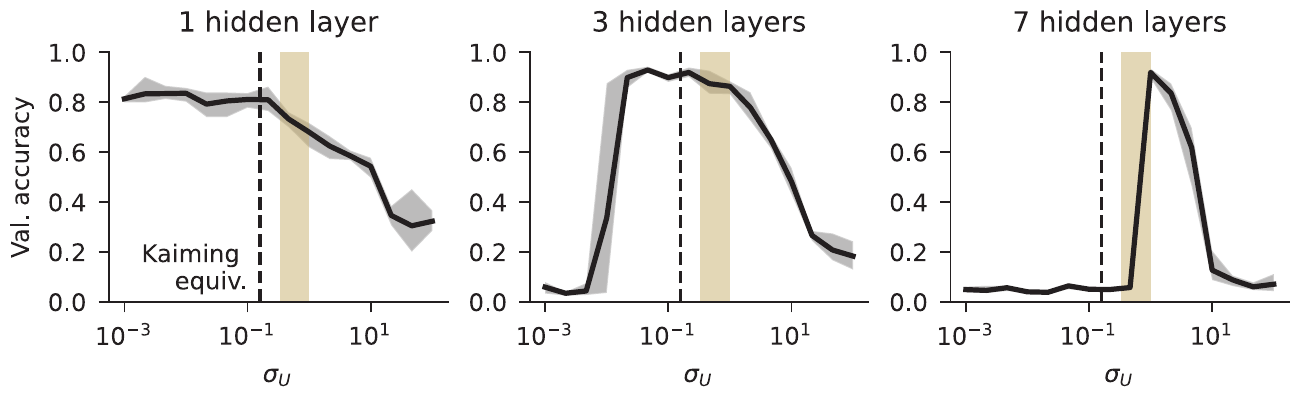}
\caption{
\textbf{Deep feed-forward \acp{CSNN} are sensitive to initialization.}
Validation accuracy as a function of target membrane potential fluctuation strength $\sigma_U$ for strictly feed-forward CSNNs of increasing depth. All networks were trained on the SHD dataset. The shaded region around the lines indicates the range of values across five random seeds. The sand-colored shaded region corresponds to our suggested target fluctuation magnitude $ \frac{1}{3}\le \sigma_U\le 1$. The dashed line indicated Kaiming initialization.
}
\label{sfig:deep-forward}
\end{figure}

\begin{figure}[htb]
	\includegraphics[width=1\textwidth]{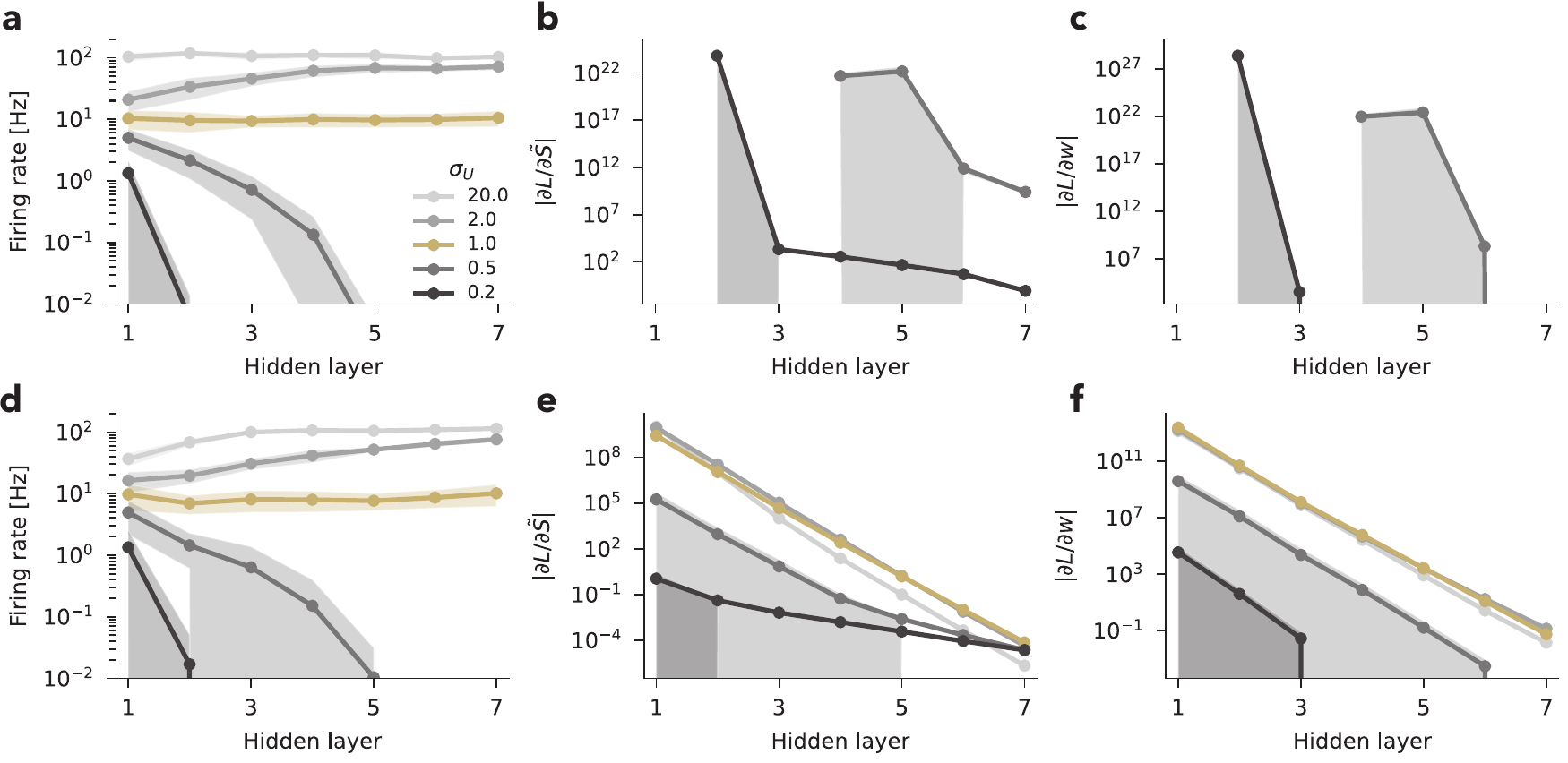}
\caption{
\textbf{Re-scaled surrogate gradients can prevent vanishing gradients
at the cost of exploding gradients.}
\textbf{(a)} Population firing rate at time of initialization  as a function of hidden layers in a \ac{CSNN} with seven hidden layers, for different target fluctuation magnitudes $\sigma_U$. The shaded region around the lines indicates one standard deviation across five random seeds.
\textbf{(b)} The magnitude of surrogate gradients $\nicefrac{\partial L}{\partial \tilde{S}}$ as a function of hidden layer. The network is being trained with a re-scaled version of the SuperSpike nonlinearity that ensures propagation of gradients even in the absence of spikes (see Methods). For initializations with $\sigma_U\geq1$, the gradients explode.
\textbf{(c)} As panel (a), but displaying the magnitude of weight changes $|\nicefrac{\partial L}{\partial W}|$. When neurons in the previous layer are quiescent, the weight update equals zero.
\textbf{(d)-(f)} As panels (a)-(c), for a \ac{CSNN} without recurrent connections in hidden layers.
}
\label{sfig:surrgrad-rescaled}
\end{figure}

\begin{figure}[htb]
	\includegraphics{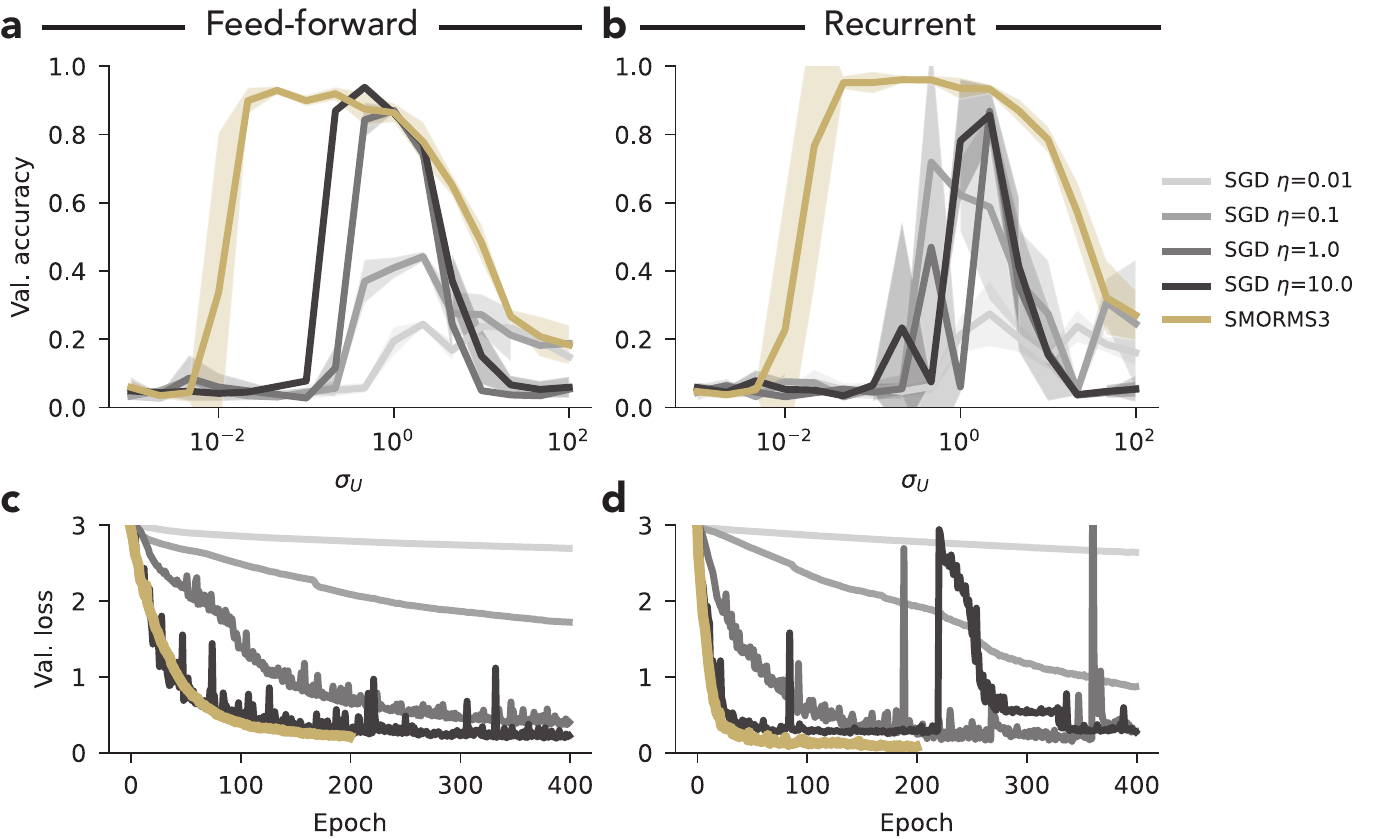}
\caption{
\textbf{Robustness to weight initialization is sensitive to the choice of optimizer.}
	\textbf{(a)} Validation accuracy after training as a function of the target fluctuation magnitude $\sigma_U$ at initialization for feed-forward \acp{CSNN} with three hidden layers trained on the SHD dataset. Networks were either trained with \ac{SGD} with learning rate $\eta$ or with the SMORMS3 optimizer (see Methods). The shaded region around the lines indicates one standard deviation across five random seeds.
	\textbf{(b)} As Panel (a), for recurrently connected \acp{CSNN} with three hidden layers.
	\textbf{(c)} Validation loss as a function of training epochs for the best performing feed-forward networks of each optimization scheme plotted in panel (a).
	\textbf{(d)} As panel (c), for the recurrently connected \acp{CSNN} plotted in panel (b).
}
\label{sfig:optimizers}
\end{figure}

\begin{figure}[htb]
	\includegraphics[width=1\textwidth]{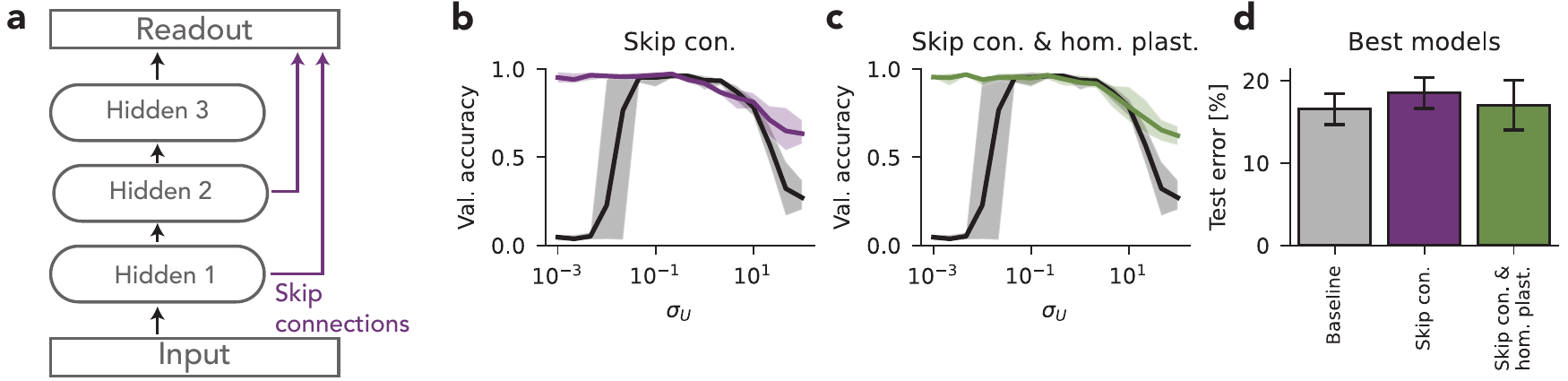}
\caption{\textbf{Skip connections can increase robustness to initialization in deep \acp{CSNN}.}
\textbf{(a)} Illustration of skip connections in deep \acp{CSNN}. Readout units receive inputs from every hidden layer separately.
\textbf{(b)} Validation accuracy after training \acp{CSNN}  with three hidden layers either with (colored line) or without skip connections on the SHD dataset. The shaded region around the lines indicates the range of values across five random seeds. 
\textbf{(c)} As panel (b), but the colored line corresponds to networks with both skip connections and homeostatic plasticity.
\textbf{(d)} Test accuracy of the 5 best performing models in terms of validation accuracy for models trained with skip connections, combined skip connections and homeostatic plasticity and the baseline model. 
}
\label{sfig:skipcon}
\end{figure}

\begin{figure}[htb]
	\includegraphics[width=1\textwidth]{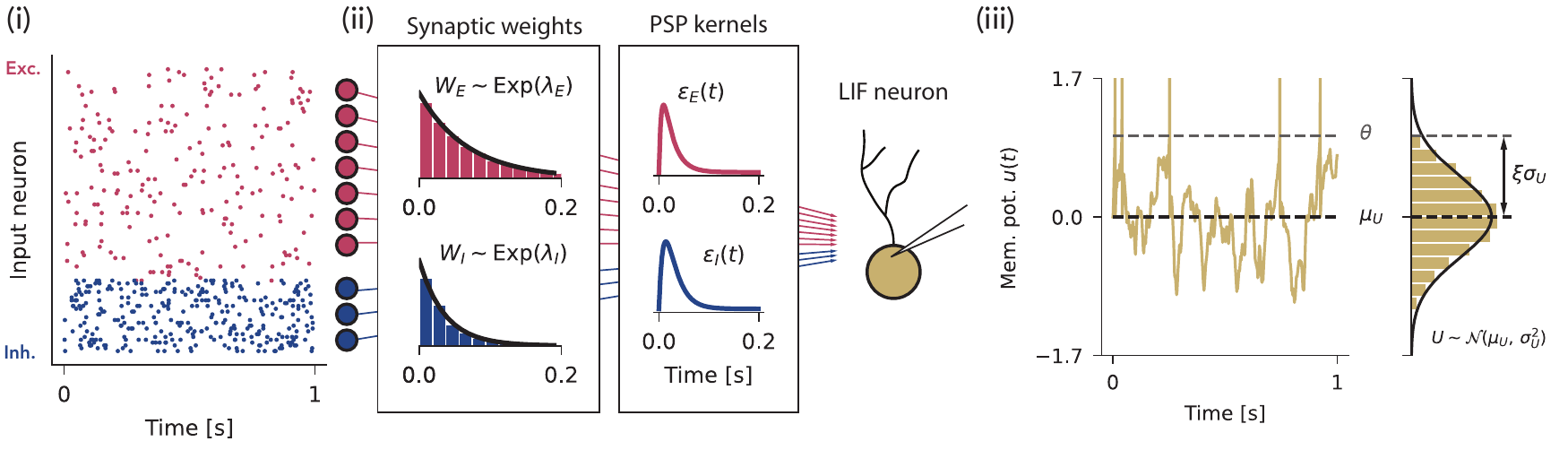}
\caption{\textbf{Parameterization of fluctuation-driven spiking as an initialization strategy for Dalian \acp{SNN}.}
	Incoming presynaptic Poisson spike trains (i) from separate excitatory (red) and inhibitory (blue) populations are weighted by the respective synaptic strengths $W^E$ and $W^I$ and filtered through the respective \ac{PSP} kernels $\epsilon_E(t)$ and $\epsilon_I(t)$ (ii) to yield membrane potential fluctuations $u(t)$ in a postsynaptic neuron (iii). As in the non-Dalian case, the magnitude of membrane potential fluctuations, $\sigma_U$, is determined by the parameters of the presynaptic weight distributions, $\lambda^E$ and $\lambda^I$. Synaptic weights can thus be initialized from a target $\sigma_U$ (see Methods).
}
\label{sfig:dale-theory}
\end{figure} 
\clearpage
\section*{Supplementary Tables}

\begin{table}[tpbh]
\setlength{\tabcolsep}{5pt}
\centering
\begin{tabular*}{\textwidth}{@{\extracolsep{\fill}}lcccc}
\hline
 	            					    & Randman 	& SHD 		& CIFAR-10 		& DVS-Gestures 	\\
\hline 
 Duration $T_\text{dataset}$ [ms]   	& 200		& 700		& 100			& 1,000			\\
 Input dimensions		                & 1			& 1			& 2				& 3				\\
 Input neurons $n_\text{dataset}$	    & 20		& 700		& 32 x 32		& 2 x 32 x 32	\\
 Firing rate $\nu_\text{dataset}$ [Hz]  & 5         & 15.8      & 14.3          & 9.2           \\
 Classes		                        & 10		& 20		& 10			& 11			\\
 Total training samples	                & 8,000		& 7,340		& 54,000		& 1,077			\\
\hline
\end{tabular*}
\caption{Dataset specifications after pre-processing.}
\label{stab:datasets}
\end{table}

\begin{table}[htpb]
\setlength{\tabcolsep}{5pt}
\centering
\begin{tabular*}{\textwidth}{@{\extracolsep{\fill}}lcc}
\toprule
 								        & Randman		                    & SHD 		\\
\midrule 
Nb. of input neurons			        & 20				                    & 700		\\
Nb. of output neurons			    & 10				                    	& 20			\\
Nb. of hidden neurons			    & 128                      			& 128 or 160		\\
Nb. of hidden neurons (Dalian \acp{SNN})		& -		        				& 128 exc. / 32 inh.			\\
Minibatch size					        & 400                   				& 400			\\
Nb. of training epochs				    & 200	                    		& 200			\\
\bottomrule
\end{tabular*}
\caption{Network and training parameters used for simulations of fully connected \acp{SNN} with a single hidden layer on the Randman and SHD datasets.}
\label{stab:architecture_shallow}
\end{table}

\begin{table}[htpb]
\setlength{\tabcolsep}{5pt}
\centering
\begin{tabular*}{\textwidth}{@{\extracolsep{\fill}}lccc}
\toprule
 								& SHD 			& CIFAR-10 				& DVS-Gestures 				\\
\midrule 
Nb. of input neurons			& 700			& $32 \times 32$		    & $2 \times 32 \times 32$   \\
Nb. of output neurons		& 20				& 10	                    	& 11			\\
Nb. of hidden layers			& 1, 3 or 7			& 2 or 4					& 6 or 8						\\
Nb. of hidden neurons 		& 16-32-64-64-64-64-64 & 32-32-64-64		& 32-32-64-64-128-128-128-128 \\
Minibatch size					& 100-400		& 128					& 32						\\
Nb. of training epochs		& 200			& 50					    & 20						\\
\bottomrule
\end{tabular*}
\caption{Network and training parameters used for simulations of deep convolutional \acp{SNN} on the SHD, CIFAR-10 and DVS-Gestures datasets. The hidden layer sizes of architectures with $m$ hidden layers are given as the first $m$ values in the row "Nb. of hidden neurons".}
\label{stab:architecture_CSNN}
\end{table}

\begin{sidewaystable}[htpb]
\def\arraystretch{1.4}\setlength{\tabcolsep}{5pt}
\caption{Summary of strategies for fluctuation-driven initialization of Dalian \acp{SNN}.}
\centering
\begin{tabularx}{\textheight}{@{\extracolsep{\fill}}lllll}
\toprule
\makecell{\textbf{Network} \\ \textbf{Architecture}}
& \makecell{\textbf{Weight} \\ \textbf{Distribution}} 	
& \textbf{Weight Parameters}
& \textbf{}
& \makecell{\textbf{Good regime for} \\ \textbf{Initialization}} 
 \\
\midrule
\addlinespace
Feed-forward 
& $\begin{aligned}
        W_E &\sim \mathrm{Exp}\left(\lambda_E\right)\\
        W_I &\sim \mathrm{Exp}\left(\lambda_I\right)
   \end{aligned}$
& $\begin{aligned} 
\lambda_E &= \frac{ \sqrt{2(\Delta_{EI}^2 n_E\nu_E\hat\epsilon_E +
n_I\nu_I\hat\epsilon_I)}}{\sigma_U\Delta_{EI}}\\
           \lambda_I &= \Delta_{EI}\lambda_E\\
    \end{aligned}$
& $\Delta_{EI} = \frac{n_I \nu_I \bar\epsilon_I}{n_E \nu_E \bar\epsilon_E}$
& $\begin{aligned}
    \mu_U&=0\\ 
    \frac{1}{3}\le\sigma_U&\le 1\\ 
\end{aligned}$
\\

\addlinespace
\midrule
\addlinespace
Recurrent
& $\begin{aligned}
        W_F &\sim \mathrm{Exp}\left(\lambda_F\right)\\
        W_R &\sim \mathrm{Exp}\left(\lambda_R\right)\\
        W_I &\sim \mathrm{Exp}\left(\lambda_I\right)
   \end{aligned}$
& $\begin{aligned} 
            \lambda_F &=  \frac{ \sqrt{2 \nu \left((\Delta_{EI}^{R})^2 \hat\epsilon_E N_R + \Delta_R^2\left(\Delta_{EI}^R N_F \hat\epsilon_E + N_I \hat\epsilon_I \right) \right)} }{ \sigma_U \Delta_{R} \Delta_{EI}^R}\\ 
           \lambda_R &= \lambda_F\Delta_R \\
           \lambda_I &= \lambda_F \Delta_{EI}^R
    \end{aligned}$
& $ \begin{aligned}
    \Delta_{R} &= \sqrt{ \frac{\alpha N_R}{N_F - \alpha N_F} }\\
    \Delta_{EI}^R &= \frac{ \Delta_R \bar\epsilon_I N_I }{\Delta_R \bar\epsilon_E N_F + \bar\epsilon_E N_R}
\end{aligned}$
& $\begin{aligned}
    \mu_U&=0\\ 
    \frac{1}{3}\le\sigma_U&\le 1\\ 
    0<\alpha&<1
\end{aligned}$
\\
\addlinespace
\bottomrule
\end{tabularx}
\label{stab:dalian_init}
\end{sidewaystable}
 
\clearpage
\section*{Supplementary Material}

\subsection{PSP kernel parameters} 
\label{sup:kernel}

\paragraph{Current-based synapses.} 
The parameters $\bar\epsilon$ and
$\hat\epsilon$ characterize the integral of the \ac{PSP} kernel $\epsilon(t)$, and the integral of the squared \ac{PSP} kernel $\epsilon(t)^2$, respectively:
\begin{eqnarray}    
    \bar\epsilon &=& \int_{-\infty}^\infty \epsilon(s) ds 
    \label{eq:ebar}\\
    \hat\epsilon &=& \int_{-\infty}^\infty \epsilon(s)^2 ds~.
    \label{eq:ehat}
\end{eqnarray} 
To arrive at analytical expressions for $\bar\epsilon$ and
$\hat\epsilon$, we first derive the kernel $\epsilon(t)$ from the differential equations of a \ac{LIF} neuron \citep{Gerstner2002}
\begin{eqnarray}
\tau_{\text{mem}} \frac{du(t)}{dt} &=& -u(t) + I(t) 
\label{eq:lif-membrane} 
\\
\frac{dI(t)}{dt} &=& -\frac{I(t)}{\tau_{\text{syn}}} + \sum_j w_jS_j(t)~,
\label{eq:lif-current}
\end{eqnarray}
where $\tau_{\text{mem}}$ and $\tau_{\text{syn}}$ are the membrane and synaptic time constants, $S_{j}(t)$ are the input spike trains from the presynaptic neuron $j$ weighted by the synaptic weight $w_j$, $u(t)$ is the membrane potential and $I(t)$ is the current.\\
To obtain an expression for the kernel $\epsilon(t)$, we consider that neuron
$i$ receives a single spike from one presynaptic neuron $j$ at time $t=0$ with
a synaptic weight of $w_j=1$. In this case, the synaptic current is simply
\begin{equation}
    I(t) = \exp\left(-\frac{t}{\tau_{\text{syn}}}\right),
\end{equation}
which can be inserted into equation \eqref{eq:lif-membrane} to obtain an
explicit solution for the membrane potential
\begin{equation} \label{eq:kernel}
u(t) = \frac{1}{1-\frac{\tau_{\text{mem}}}{\tau_{\text{syn}}}}
\left(\exp\left(-\frac{t}{\tau_{\text{syn}}}\right)-
\exp\left(-\frac{t}{\tau_{\text{mem}}}\right)\right) \Theta(t)~.
\end{equation}
Eq. \eqref{eq:kernel} corresponds to the \ac{PSP}-kernel $\epsilon(t)$ evoked
by a single presynaptic spike, provided that the membrane potential stays in
the sub-threshold regime. Note that in the limit of
$\tau_{\text{mem}}\rightarrow \tau_{\text{syn}}$, the membrane potential
follows a scaled alpha function
\begin{equation}
\lim_{\tau_{\text{mem}}\to\tau_{\text{syn}}} u(t) = \frac{t}{\tau_{\text{syn}}}
\exp\left(-\frac{t}{\tau_{\text{mem}}}\right)\Theta(t)~.
\end{equation}
We can now solve the integrals in Eqs.~\eqref{eq:ebar} and
\eqref{eq:ehat} that define the kernel parameters $\bar\epsilon$ and
$\hat\epsilon$, starting with $\tau_{\text{mem}}\ne\tau_{\text{syn}}$:
\begin{eqnarray}
\bar\epsilon &=& \int_{-\infty}^{\infty}\epsilon(t)dt 
\nonumber 
\\
&=&  \int_{0}^{\infty}\frac{1}{1-\frac{\tau_{\text{mem}}}{\tau_{\text{syn}}}}
\left(\exp(-\frac{t}{\tau_{\text{syn}}})-
\exp(-\frac{t}{\tau_{\text{mem}}})\right)dt 
\nonumber 
\\
&=&\frac{1}{1-\frac{\tau_{\text{mem}}}{\tau_{\text{syn}}}}\left(-\tau_{\text
{syn}}\exp\left(-\frac{t}{\tau_{\text{syn}}}\right)+\tau_{\text{mem}}\exp\left(
-\frac{t}{\tau_{\text{mem}}}\right)\right)\Bigg|_0^\infty 
\nonumber 
\\
&=&\tau_{\text{syn}}
\label{eq:ebar-sol}
\end{eqnarray}

{\small
\begin{eqnarray}
\hat\epsilon &=& \int_{-\infty}^{\infty} \epsilon^2(t)dt 
\nonumber 
\\
&=&\int_{0}^{\infty}\left[\frac{1}{1-\frac{\tau_{\text{mem}}}{\tau_{\text{syn}}
}}
\left(\exp\left(-\frac{t}{\tau_{\text{syn}}}\right) -
\exp\left(-\frac{t}{\tau_{\text{mem}}}\right)\right)\right]^2dt 
\nonumber 
\\
&=& \left(\frac{1}{1-\frac{\tau_{\text{mem}}}{\tau_{\text{syn}}}}\right)^2
\Bigg(\frac{2\tau_{\text{mem}}\tau_{\text{syn}}\exp\left(-\frac{t}{\tau_{\text
{syn}}}-\frac{t}{\tau_{\text{mem}}}\right)}{\tau_{\text{syn}}+\tau_{\text{mem}}
}
 \frac{\tau_{\text{syn}}\exp\left(-2\frac{t}{\tau_{\text{syn}}}\right)}{2}
-\frac{\tau_{\text{mem}}\exp\left(-2\frac{t}{\tau_{\text{mem}}}\right)}{2}\Bigg
) \Bigg|_0^\infty 
\nonumber 
\\
&=& \frac{\tau_{\text{syn}}^2}{2(\tau_{\text{syn}}+\tau_{\text{mem}})}~.
\label{eq:ehat-sol}
\end{eqnarray}
}

When solving for $\bar\epsilon$ and $\hat\epsilon$ in the case of $\tau_{\text{mem}} = \tau_{\text{syn}}$ (taking the $\lim_{\tau_{\text{mem}}\to \tau_{\text{syn}}}$ and applying de L’Hospital's rule), one will arrive at the same solutions as in the above case (see Tab.~\ref{stab:epsilon}).

\paragraph{Delta-Synapses.} For reasons of simplicity, current-based synaptic
transmission is often replaced with 'delta synapses' in \acp{SNN}. 
In this case, the membrane potential dynamics in response to a single input
spike can be described as a mono-exponential decay
\begin{equation}
u(t) = \exp\left(-\frac{t}{\tau_{\text{mem}}}\right)
\end{equation}
and the kernel parameters simplify to
\begin{eqnarray}
\bar\epsilon &=& \int_{-\infty}^{\infty}\epsilon(t)dt = \int_{0}^{\infty}
\exp\left(-\frac{t}{\tau_{\text{mem}}}\right) dt = \tau_{\text{mem}}
\\
\hat\epsilon &=& \int_{-\infty}^{\infty}\epsilon^2(t)dt =
\int_{0}^{\infty}\left[ \exp\left(-\frac{t}{\tau_{\text{mem}}}\right)\right]^2
dt = \frac{\tau_{\text{mem}}}{2}~.
\end{eqnarray}
A summary of analytical expressions for $\bar\epsilon$ and $\hat\epsilon$ can be found in Tab.~\ref{stab:epsilon}.

\begin{figure}[htb]
	\includegraphics[width=0.45\textwidth]{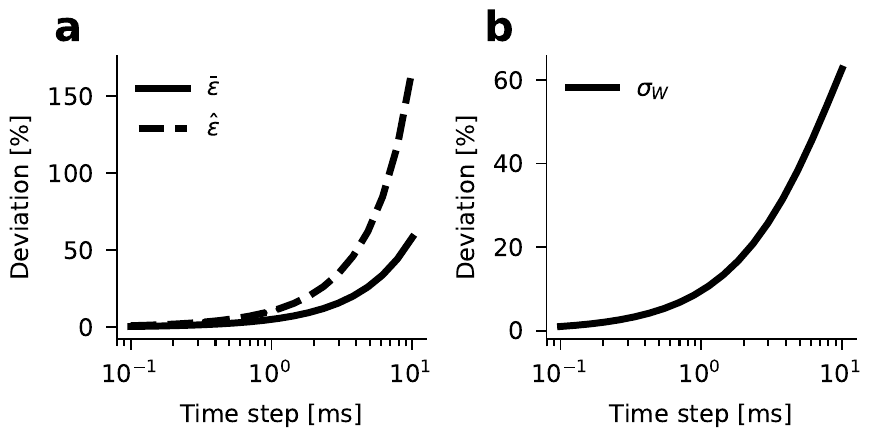}
\caption{\textbf{Numerical calculation of $\bar\epsilon$ and $\hat\epsilon$.}
	\textbf{(a)} Difference (in \% of the analytical solution) between numerically calculated and analytically calculated $\bar\epsilon$ and $\hat\epsilon$ as a function of the simulation time step. 
	\textbf{(b)} The resulting difference in the standard deviation of synaptic weights $\sigma_W$ using fluctuation-driven initialization.
}
\label{sfig:kernel}
\end{figure}

\paragraph{Numerical estimation of $\bar\epsilon$ and $\hat\epsilon$.}
Note that the analytical solutions summarized in Tab.~\ref{stab:epsilon}
might not reflect the values of $\bar\epsilon$ and $\hat\epsilon$ obtained in numerical simulations
of the \acp{SNN} employing the neuronal dynamics. 
Specifically, numerical simulations of \acp{SNN} often operate with a large
time step and integrate neuronal dynamics using simple forward Euler
approaches. 
A more accurate approach with regards to the fluctuations in numerical
simulations would therefore be to solve the integrals using the same numerical
approximation as employed during the simulations.
To quantify the degree to which the analytical solutions for $\bar\epsilon$ and $\hat\epsilon$ 
deviate from the numerical approximations of the integrals, we numerically approximated the integrals
with different values for the simulation time step.
Indeed, if the simulation time step was large ($\geq 1$\,ms), the numerical solution substantially
deviated from the analytical solution (Fig.~\ref{sfig:kernel}).
Hence, for all numerical simulations in this paper, we calculated $\bar\epsilon$ and $\hat\epsilon$ through numerical approximation (forward Euler) using the same time step as during \ac{SNN} training. The numerical values for $\bar\epsilon$ and $\hat\epsilon$ used throughout our numerical simulations can be found in Tab.~\ref{tab:epsilon-numerical} of the main text.

\renewcommand{\arraystretch}{1.5}
\begin{table}[htpb]
\setlength{\tabcolsep}{5pt}
\centering
\begin{tabular*}{0.6\textwidth}{@{\extracolsep{\fill}}ccc}
\toprule
 								& Delta synapses 						& Current-based synapses  \\
\midrule
$\bar \epsilon$					& $\tau_{\text{mem}}$					& $\tau_{\text{syn}}$				      \\
$\hat \epsilon$					& $\frac{\tau_{\text{mem}}}{2}$			&
$\frac{\tau_{\text{syn}}^2}{2(\tau_{\text{syn}}+\tau_{\text{mem}})}	$
\\
\bottomrule
\end{tabular*}
\caption{Analytical expressions for $\bar\epsilon$ and
$\hat\epsilon$ for LIF neurons with delta synapses or current-based synapses.}
\label{stab:epsilon}
\end{table}
\renewcommand{\arraystretch}{1.0}

\clearpage

\subsection{Population-level variability in membrane potential fluctuations}
\label{sup:popvar}

The initialization strategy discussed in this article is based on target
values for the membrane potential mean $\mu_U$ and its standard deviation
$\sigma_U$.
Due to the inherent stochasticity arising from random sampling of synaptic
weights, we can expect deviations from these targets in the numerically observed membrane potential with mean $\hat \mu_U$ and $\hat \sigma_U$. 
As illustrated in Figs.~\ref{fig:valid} and~\ref{sfig:shallow} in the main text, this variability can cause some neurons to fire in the mean-driven regime, even if the initialization targets
are set in the fluctuation-driven regime. 

Here, we analyze the expected variability of membrane potential fluctuations across a 
population of $m$ postsynaptic neurons with independently drawn weight vectors
$\{\vec{w}_{1}, \vec{w}_{2}, ... \vec{w}_{m}\}$. Specifically, we are interested in the 
sampling distributions of $\hat \mu_U$ and $\hat \sigma_U^2$. 
For simplicity, we ignore the spiking dynamics of the \ac{LIF} neuron model and assume that synaptic weights are independently drawn from the zero-mean Normal distribution $W \sim \mathcal{N}(0,\,\sigma_W^2)$. 

\paragraph{Sampling distribution of $\mu_W$ and $\sigma_W$.} 
In order to derive the sampling distributions of the membrane potentials, we first need to derive the sampling distribution of the synaptic weights. 
That is, for independently drawn weight vectors $\{\vec{w}_{1}, \vec{w}_{2}, ... \vec{w}_{m}\}$ of size $n$, we are interested in the distributions of the sample mean $\hat \mu_W$ and the sample variance $\hat \sigma_W$. For weights drawn from the zero-mean distribution $W \sim \mathcal{N}(0,\,\sigma_W^2)$, the former is simply
\begin{equation}
    \hat \mu_W \sim \mathcal{N} \left(0, \frac{\sigma_W^2}{n} \right) ~.
\end{equation}
To obtain the sampling distribution of the variance, we first observe that, according to Cochran's Theorem \citep{Cochran1934-rw}, 
\begin{equation}
    n \frac{\hat \sigma_W^2}{\sigma_W^2} \sim \chi^2_{n-1}  ~,
\end{equation}
which we can alternatively express as a special case of a Gamma distribution $\Gamma\left(k, \theta \right)$ with shape parameter $k=\frac{n-1}{2}$ and scale parameter $\theta=2$. Hence the above equation can be written as
\begin{equation}
    n \frac{\hat \sigma_W^2}{\sigma_W^2} \sim \Gamma \left(\frac{n-1}{2}, 2 \right) ~ .
    \label{eq:gamma_dist_var-1}
\end{equation}
Using the scaling property of the gamma function, we can solve Eq.~\eqref{eq:gamma_dist_var-1} for $\hat \sigma_W^2$ under the requirement $\frac{\sigma_W^2}{n}>0$. This holds as long as we have variance in the weight initialization. Hence we can express the distribution of the sample variance $\hat\sigma_W^2$ as
\begin{equation}
    \hat \sigma_W^2 \sim \Gamma \left(\frac{n-1}{2}, \frac{2 \sigma_W^2}{n} \right)  ~.
\end{equation}

\paragraph{Sampling distribution of $\hat \mu_U$.}
We start by observing that the membrane potential $U$ is normally distributed according to the Central Limit Theorem and its mean $\mu_U$ and variance $\sigma_U^2$ were given in Eqs.~\eqref{eq:mu_u} and~\eqref{eq:sigma_u} in the main text (see Fig.~\ref{fig:theory}). 
To derive the sampling distribution of $\hat \mu_U$, we observe that the sample $\hat \mu_U$ is related to $\hat \mu_W$ through
\begin{equation}
    \hat \mu_U = n \nu \bar \epsilon \hat \mu_W  ~.
\end{equation}
From the above equation and our derivation of the sampling distribution of $\hat \mu_W$, it becomes apparent that
\begin{equation}
    \hat \mu_U \sim \mathcal{N} \left(0, n^2 \nu^2 \bar \epsilon^2 \left(\frac{\sigma_W^2}{n}\right) \right)  ~,
\end{equation}
which can be further simplified to 
\begin{equation}
    \hat \mu_U \sim \mathcal{N} \left(0, \sigma_U^2 \frac{\nu \bar \epsilon^2}{\hat\epsilon} \right)
\end{equation}
by expressing it in terms of the initialization target $\sigma_U$, for which we inserted Eq.~\eqref{eq:wsigma-simple} from the main text (Fig. \ref{sfig:popvar}a-d).

We can conclude that random sampling of the weights induces systematic variance in the expected membrane potential means $\hat \mu_U$ of neurons receiving inputs from $n$ homogeneous Poisson processes with firing rate $\nu$. Specifically, the expected variance on the population level is independent of the number of inputs $n$, but scales with the target fluctuation magnitude $\sigma_U$ and the input firing rate $\nu$.

\paragraph{Sampling distribution of $\hat \sigma_U$.}
We follow a similar approach to derive the sampling distribution of fluctuation magnitudes in the population. Formally, we are looking for the distribution of $\hat \sigma_U^2$, which can be derived from observing that for a neuron $i$
\begin{equation} \label{eq:hat_sigma_u_i}
    \left(\hat \sigma_U^{(i)}\right)^2 = n \left(\left(\hat \sigma_W^{(i)}\right)^2 + \left(\hat \mu_W^{(i)} \right)^2 \right) \nu \hat \epsilon  ~.
\end{equation}
To get the distribution of $\hat \sigma_U$, we first need to determine the distribution of the right hand side. Starting with the distribution of $\hat\mu_W^2$, we observe that the standardized form of $\hat\mu_W^2$ follows a chi-square distribution with one degree of freedom:
\begin{equation}
    \frac{\hat \mu_W^2}{\frac{\sigma_W^2}{n}} \sim \chi^2_1 ~.
\end{equation}
Similarly to the first paragraph, we can rewrite the chi-square distribution as a Gamma distribution and use its scaling properties to obtain the distribution of $\hat \mu_W^2$:
\begin{equation}
    \hat \mu_W^2 \sim \Gamma\left(\frac{1}{2}, \frac{2 \sigma_W^2}{n} \right) ~ .
\end{equation}
Note that both $\hat \mu_W^2$ and $\hat \sigma_W^2$ are Gamma distributed with a shared scale parameter $\theta = \frac{2 \sigma_W^2}{n}$ and that these random variable are independent. We can therefore use the summation property of the Gamma distribution to obtain
\begin{equation}
    \left(\hat \sigma_W^2 + \hat \mu_W^2 \right) \sim \Gamma \left(\frac{n}{2}, \frac{2 \sigma_W^2}{n} \right)
\end{equation}
and finally plug this result back into Eq.~\eqref{eq:hat_sigma_u_i} to obtain the distribution
\begin{equation}
    \hat \sigma_U^2 \sim \Gamma \left(\frac{n}{2}, 2 \nu \hat\epsilon \sigma_W^2 \right) ~ .
\end{equation}
As we did in the previous paragraph, we can insert the solution for $\sigma_W^2$ in the case of centered weights, given by Eq.~\eqref{eq:wsigma-simple}, to simplify the distribution to
\begin{equation}
    \hat \sigma_U^2 \sim \Gamma \left(\frac{n}{2}, \frac{2 \sigma_U^2}{n} \right) 
\end{equation}
as displayed in Fig.~\ref{sfig:popvar}f.
We can alternatively express the expected variability as the distribution of standard deviations, which follows the Nakagami distribution \citep{Huang2016}
\begin{equation}
    \hat \sigma_U \sim \mathrm{Nakagami}\left(\frac{n}{2}, \sigma_U^2 \right)  
\end{equation}
with shape parameter $m=\frac{n}{2}$ and spread parameter $\Omega=\sigma_U^2$ (Fig.~\ref{sfig:popvar}f). 
Thus, random sampling of synaptic weights induces a systematic variance in $\sigma_U$ that scales with $\sigma_U^2$ and inversely with the number of inputs.

\clearpage

\subsection{Fluctuation-driven initialization of Dalian networks using log-normally distributed weights}
\label{sec:lognormal}

The initialization of the weights in separate excitatory and inhibitory neuronal populations, as is the case with Dalian networks, requires weights sampled from one-sided distributions. 
Inspired by neurobiological evidence \citep{buzsaki_log-dynamic_2014}, we consider weights sampled from the log-normal distribution parameterized by $\mu$ and $\sigma>0$ that gives rise to a random variable $X = e^{\mu + \sigma Z}$ where $Z \sim N(0, 1)$ is a standard normal random variable. The expected value and the variance of $X$ are then defined as
\begin{eqnarray}
    \mathbb{E}[X] &=& \exp\left(\mu + \frac{\sigma^2}{2}\right)\\
    \mathbb{V}[X] &=& \left(\exp\left(\sigma^2 \right) - 1 \right) \exp\left(2\mu+\sigma^2\right)~.
\end{eqnarray}
    
To parameterize synaptic weights, we use the log-normal distribution to obtain excitatory weights $\ln(W^E) \sim \mathcal{N}(\mu_E, \sigma_E)$ and inhibitory weights $\ln(W^I) \sim \mathcal{N}(\mu_I, \sigma_I)$. As we only have two equations, which we use to describe the membrane potential of a neuron in the fluctuation-driven regime (mean and variance of the membrane potential), we also need to restrict the parameterization to two parameters in total. Hence, we set $\sigma_E = \sigma_I = 1$. We can therefore simplify the above to
\begin{eqnarray}
    \mathbb{E}[W] &=& e^{\mu + 1/2} \\
    \mathbb{V}[W] &=& e^{2 \mu + 2} - e^{2\mu + 1}
\end{eqnarray}
and also note that the second moment of the distribution is
\begin{equation}
    \mathbb{E}[W^2] = e^{2 \mu + 2} ~.
\end{equation}
Given these definitions, we can write down the values of $\mu_U$ and $\sigma_U^2$ as
\begin{eqnarray}
    \mu_U &=& N_E\nu_E\bar\epsilon_Ee^{(\mu_E + 1/2)} - N_I\nu_I\bar\epsilon_Ie^{(\mu_I + 1/2)}
    \label{eq:ln_mean_u}\\
    \sigma^2_U &=& N_E\nu_E\hat\epsilon_E e^{(2\mu_E + 2)} + N_I\nu_I\hat\epsilon_I e^{(2\mu_I + 2)}  ~.
    \label{eq:ln_var_u}
\end{eqnarray}
We once more set a target mean membrane potential $\mu_U = 0$ to achieve a balanced state at initialization. Using this, we solve Eq.~\eqref{eq:ln_mean_u} to receive an expression for $\mu_I$

\begin{equation}
    \mu_I = \mu_E + \log \left(\frac{N_E\nu_E\bar\epsilon_E}{N_I\nu_I\bar\epsilon_I} \right)~ ,
\end{equation}
which can be further simplified to 
\begin{equation}
    \mu_I = \mu_E + \log \left(\frac{1}{\Delta_{EI}}\right)
    \label{eq:ln_nd_mu_i}
\end{equation}
by using the definition of $\Delta_{EI}$ from Eq.~\eqref{eq:nd_delta_ei}.
Substituting this result into Eq.~\eqref{eq:ln_var_u} allows us to solve for $\mu_E$ and we receive
\begin{equation}
    \mu_E = \frac{1}{2} \log\left(\frac{\sigma_U^2}{N_E \nu_E \hat\epsilon_E + N_I \nu_I \hat\epsilon_I \left(\frac{1}{\Delta_{EI}} \right)^2 } \right) - 1 ~.
    \label{eq:ln_nd_mu_e}
\end{equation}

Finally, Equations~\eqref{eq:ln_nd_mu_i} and \eqref{eq:ln_nd_mu_e} together with $\sigma_E=\sigma_I = 1$ and $\mu_U=0$ provide us the parameters to initialize the inhibitory and excitatory weights sampled from a log normal distribution.
\paragraph{Dalian networks with excitatory recurrence.}
We start from the mean and variance of the membrane potential for a neuron receiving a feed-forward excitatory, recurrent excitatory and recurrent inhibitory input, where we assume $\nu_R=\nu_F=\nu_I=\nu$,
\begin{eqnarray}
    \mu_U &=& \left(N_{F}\nu\bar\epsilon_{E}\right)e^{\mu_F + 1/2} + \left(N_{R}\nu\bar\epsilon_{E}\right)e^{\mu_R + 1/2} - \left(N_{I}\nu\bar\epsilon_{I}\right)e^{\mu_I + 1/2}\\
    \sigma^2_{U} &=&\left(N_{F}\nu\hat\epsilon_{E}\right)e^{2\mu_F + 2} + \left(N_{R}\nu\hat\epsilon_{E}\right)e^{2\mu_R + 2} + \left(N_{I}\nu\hat\epsilon_{I}\right)e^{2\mu_I + 2} 
\end{eqnarray}
and the definition of $\alpha$
\begin{eqnarray}
    \alpha &=& \frac{\text{Part of }\sigma_U^2 \text{ caused by excitatory feed-forward connections}}{\text{Part of }\sigma_U^2 \text{ caused by all excitatory connections}}\\
    &=& \frac{\left(N_{F}\nu\hat\epsilon_{E}\right)e^{2\mu_F + 2}} {\left(N_{F}\nu\hat\epsilon_{E}\right)e^{2\mu_F + 2} + \left(N_{R}\nu\hat\epsilon_{E}\right)e^{2\mu_R + 2}}  ~.
\end{eqnarray}
Here we are again requiring a mean membrane potential $\mu_U=0$ and we set the variances of the log-normal distributions to one, $\sigma_R = \sigma_F = \sigma_I = 1$.\\
Performing the same sequence of solving and substituting as in the above paragraph, we find explicit equations for the three weight distribution parameters:

\begin{eqnarray}
    \mu_R &=& \mu_F + \frac{1}{2} \log \left(N_F - \alpha N_F \right) - \log \left(\alpha N_R \right) = \mu_F + \Delta_{R}\\
    \mu_I &=& \mu_F + \frac{1}{2} \log \left( \frac{\bar\epsilon_E\left(e^{\Delta_{R}} N_R + N_F \right)}{N_I \bar\epsilon_I} \right) = \mu_F + \Delta_{EI}^R\\
    \mu_F &=& \frac{1}{2}\log \left( \frac{ \sigma_U^2 }{ \nu \left( e^{2\Delta_R} N_R \hat\epsilon_E + e^{2\Delta_{EI}^R} \hat\epsilon_I N_I + N_F \hat\epsilon_E \right) } \right) -1  ~.
\end{eqnarray}
 
\printbibliography
\end{refsection}

\end{document}